\documentclass[lettersize,journal]{IEEEtran}

\usepackage{amsmath,amsfonts}
\usepackage{algorithmic}
\usepackage{algorithm}
\usepackage{array}
\usepackage{textcomp}
\usepackage{stfloats}
\usepackage{url}
\usepackage{verbatim}
\usepackage{graphicx}
\usepackage{cite}
\usepackage{makecell}
\usepackage{subfigure}
\usepackage{balance}
\usepackage{flushend}
\usepackage{times}

\usepackage{soul}
\usepackage{url}
\usepackage[hidelinks]{hyperref}
\usepackage[utf8]{inputenc}
\usepackage[small]{caption}
\usepackage{graphicx}
\usepackage{amsmath,amsthm}
\usepackage{booktabs}
\usepackage{color}
\usepackage{amsfonts}
\usepackage{multirow}
\usepackage{pifont}
\usepackage{comment}
\usepackage{adjustbox}
\def\eg{\emph{e.g.}}

\def\ie{\emph{i.e.}}

\usepackage{graphicx}
\usepackage[margin=0.7in]{geometry}

\begin{document}

\title{Graph Neural Networks for Graphs with Heterophily: A Survey}

\author{Xin Zheng$^{\dagger}$, Yi Wang$^{\dagger}$, Yixin Liu, Ming Li$^{*}$~\emph{Member, IEEE}, Miao Zhang, Di Jin, Philip S. Yu~\textit{Fellow, IEEE}, Shirui Pan$^{*}$~\emph{Senior Member, IEEE}
\IEEEcompsocitemizethanks{
\IEEEcompsocthanksitem Xin Zheng is with the School of Computing Technologies, RMIT University, Melbourne, Australia. E-mail: xin.zheng2@rmit.edu.au.
\IEEEcompsocthanksitem Yi Wang is with the Department of Mathematics, City University of Hong Kong, Hong Kong, China. E-mail: ywan72@cityu.edu.hk.
\IEEEcompsocthanksitem Yixin Liu is with the School of Information and Communication Technology, Griffith University, Queensland, Australia. E-mail: yixin.liu@griffith.edu.au.
\IEEEcompsocthanksitem Ming Li is with Zhejiang Key Laboratory of Intelligent Education Technology and Application, Zhejiang Normal University, Jinhua, China. E-mail: mingli@zjnu.edu.cn.
\IEEEcompsocthanksitem Miao Zhang is with the School of Computer Science and Technology,  Harbin Institute of Technology (Shenzhen), Shenzhen, China. E-mail: zhangmiao@hit.edu.cn.
\IEEEcompsocthanksitem Di Jin is with the School of Computer Science
and Technology, Tianjin University, Tianjin, China. E-mail:
jindi@tju.edu.cn.
\IEEEcompsocthanksitem Philip S. Yu is with the Department of Computer Science, University of Illinois
at Chicago, Chicago, USA. E-mail: psyu@uic.edu.
\IEEEcompsocthanksitem Shirui Pan is with the School of Information and Communication Technology, Griffith University, Queensland, Australia. 
Email: s.pan@griffith.edu.au.

\IEEEcompsocthanksitem ${\dagger}$ Xin Zheng and Yi Wang contributed equally to this work. 
\IEEEcompsocthanksitem *Corresponding Authors.
}}


\markboth{Journal of \LaTeX\ Class Files,~Vol.~14, No.~8, August~2021}%
{Shell \MakeLowercase{\textit{et al.}}: A Sample Article Using IEEEtran.cls for IEEE Journals}


\maketitle

\begin{abstract}
Recent years have witnessed fast developments of graph neural networks (GNNs) that have benefited myriad graph analytic tasks and applications. Most GNNs rely on the homophily assumption that nodes belonging to the same class are more likely to be connected. However, as a ubiquitous graph property in numerous real-world scenarios, heterophily, \ie, nodes with different labels tend to be linked, significantly limits the performance of tailor-made homophilic GNNs. Hence, \textit{GNNs for heterophilic graphs} are gaining increasing research attention to enhance graph learning with heterophily. In this paper, we provide a comprehensive review of GNNs for heterophilic graphs. Specifically, we propose a systematic taxonomy that governs existing heterophilic GNN models, along with general summaries and detailed analyses. 
Furthermore, we discuss the relationship between heterophily and various graph research domains, aiming to facilitate the development of more effective GNNs across a spectrum of practical applications and learning tasks in the graph research community.
In the end, we point out potential directions to advance and inspire future research and applications on heterophilic graph learning with GNNs.

\end{abstract}
\begin{IEEEkeywords}
Graph neural networks, heterophily, graph representation learning, message passing.
\end{IEEEkeywords}

\section{Introduction}
\IEEEPARstart{G}{raphs} are pervasively structured data and have been widely used in many real-world scenarios, such as social networks~\cite{tang2009social,peng2022reinforced}, knowledge bases~\cite{vashishth2019composition}, traffic networks~\cite{wu2020connecting}, and recommendation systems~\cite{ying2018graph,ma2019learning}. Recently, graph neural networks (GNNs) have achieved remarkable success with powerful learning ability and become prevalent models to tackle various graph analytical tasks, such as node classification, link prediction, and graph classification~\cite{gcn_kipf2017semi,sage_hamilton2017inductive,vashishth2019composition,velivckovic2018graph}.

\begin{figure}[!t]
  \centering
  \includegraphics[width=0.5\textwidth]{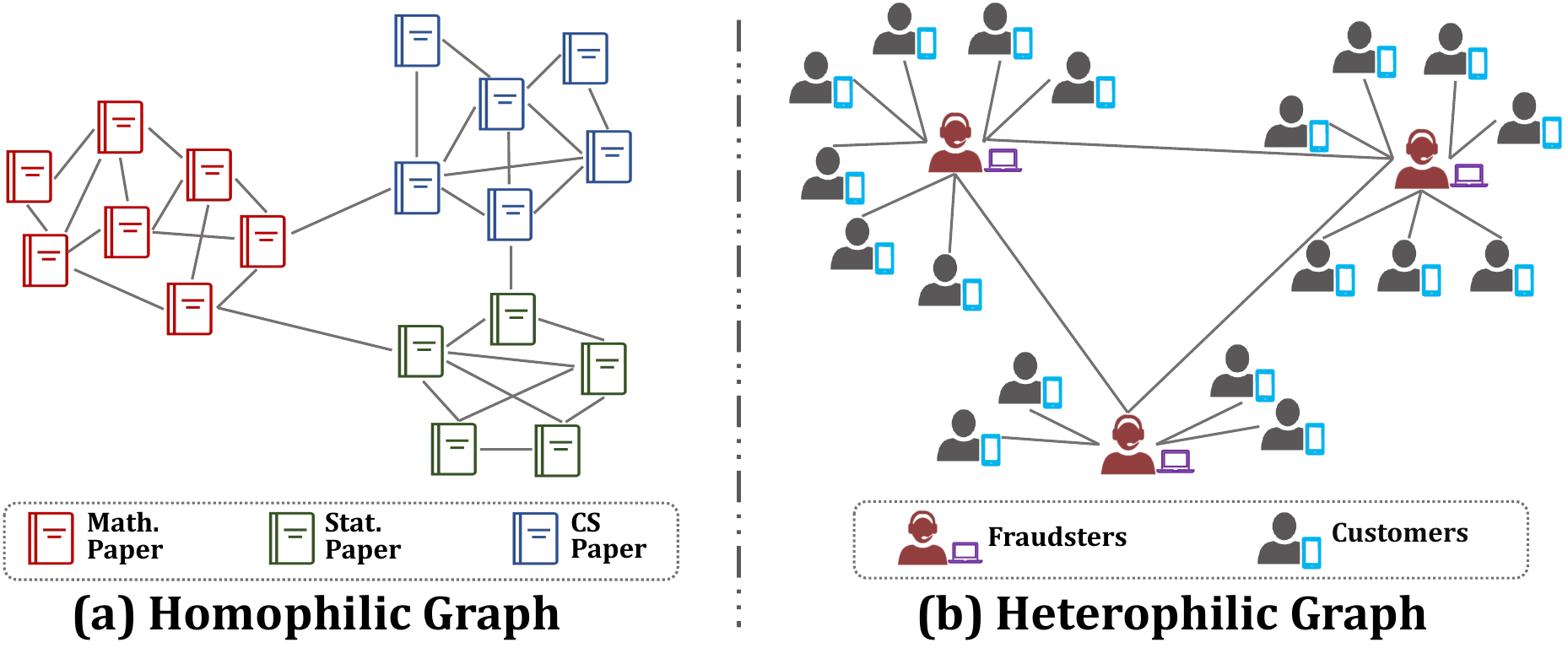}
  \caption{Examples of homophilic and heterophilic graphs (Left: (a) a citation network; Right: (b) an online transaction network).}
  \label{fig:homo_hetero}
\end{figure}

While a large number of GNNs with diverse architectures have been designed~\cite{gin_xu2019powerful,wu2020comprehensive,zhu2020graph,liu2023towards,zheng2023structure,zheng2023gnnevaluator}, the majority of them follow the homophily assumption, \ie, nodes with similar features or same class labels are linked together. For example, in citation networks, a study usually cites reference papers from the same research area~\cite{ciotti2016homophily}. However, real-world graphs do not always obey the homophily assumption but show an opposite property, \ie, {\it{heterophily}} that linked nodes have dissimilar features and different class labels~\cite{zheng2023auto,xu2022hp,liu2022towards,zheng2022multi,zheng2023finding}. For instance, in online transaction networks, fraudsters are more likely to build connections with customers instead of other fraudsters~\cite{pandit2007netprobe}; in dating networks, most people prefer to date with people of the opposite gender~\cite{zhu2021graph}; in molecular networks, protein structures are more likely composed of different types of amino acids that are linked together~\cite{h2gcn_zhu2020beyond}. The examples of homophilic and heterophilic graphs are provided in Fig.~\ref{fig:homo_hetero} to illustrate their difference visually. Importantly, such heterophily restricts the learning ability of existing homophilic GNNs on general graph-structural data, resulting in significant performance degradation on heterophilic graphs~\cite{h2gcn_zhu2020beyond,zhu2021graph,fagcn_bo2021beyond}.

\textbf{Core Challenges of GNNs for Heterophily.} 
We attribute the performance degradation to the uniform message passing framework under the homophily setting. The procedure of this framework can be summarized as: first aggregating the messages extracted from local neighbor nodes, then updating the final ego node (the current central node itself) representations with aggregated neighbor messages. 
Nevertheless, due to the heterophily property of graphs, this mechanism poses significant challenges for the development of heterophilic GNNs, primarily manifesting in two aspects:
\begin{itemize}
    \item \textbf{Challenge-1: Undiscovered Non-local Neighbors.} Guided by homophily, neighbor aggregation in homophilic GNNs restricts information extraction to the proximal local topology of graphs. When applied to graphs with heterophily, it fails to explore non-local topology, where heterophilic nodes of the same class are typically situated at long-term distances. This poses a significant challenge in identifying and learning informative nodes with high structural and semantic similarities on heterophilic graphs.
    \item \textbf{Challenge-2: Indistinguishable Node Representation Learning.} 
    Homophilic GNNs employ uniform local neighbor aggregation and update the central node representation when they are typically similar and share the same labels. Consequently, on heterophilic graphs, the discrepancy between similar non-local neighbors and dissimilar local neighbors can not be effectively captured. This results in critical challenges in learning discriminative node representations with distinguishable heterophily information through diverse customized message passing.
\end{itemize}

In light of these challenges, recently, an increasing number of researchers have started turning their attention to the study of GNNs with heterophily. The research focus is sufficiently broad, from heterophilic graph data exploration~\cite{lim2021new,lim2021large,yan2021two} to various technical algorithm development ~\cite{fagcn_bo2021beyond,yang2021diverse,chien2020adaptive,liu2022towards,zheng2023auto,zheng2023finding,liu2023beyond}.

\textbf{Importance of Developing Heterophilic GNNs.} Heterophilic graph learning with GNNs is becoming an upward trending research topic and it shows closely tied connections with diverse domains in graph research.
The compelling significance of GNN development for heterophilic graphs is underscored by the following aspects:

\begin{itemize}
    \item \textit{Enhancing the understanding of complex and diverse heterophily graphs.} Graph-structure data with the heterophilic property presents great complexity and diversity, and it is prevalent across various real-world application scenarios, ranging from daily-life personal relationships to scientific chemical molecular study. 
    A thorough and ongoing exploration of heterophily graph data would significantly enhance the understanding of complex and diverse heterophily graphs, thereby providing valuable guidance for the development of GNN models and advancements in heterophily graph learning.
    \item \textit{Advancing heterophilic graph analysis and learning.} Heterophilic graph analysis and learning tasks are still open and promising research topics in development, while numerous challenges need to be tackled for designing heterophilic GNN models with expressive performance, robustness, and generalization ability.
    The advancement of heterophilic GNNs is pivotal for unlocking the full potential of heterophilic graph analysis in addressing various practical graph learning tasks covering both heterophilic node representations and graph structures.
    
    \item \textit{Adapting with versatility in heterophilic GNN development.}
    As heterophilic graphs exist prevalently and show close connections with various graph research domains, \eg, over-smoothing and anomaly detection. Hence, developing specialized heterophilic GNN architectures and learning techniques would be pivotal in expanding the versatility and adaptability of GNN models, unleashing power of heterophilic GNNs in cross graph research domains and applications.

\end{itemize}

\textbf{Distinctions from Existing Surveys.}
This study constitutes an expansion of the domain survey initially released on arXiv (preprint 2202.07082v1), the first comprehensive review in this field. Notably, since the publication of this seminal survey in 2022, we have witnessed exponential growth in related research. Available statistics indicate that the volume of new literature generated between 2022 and 2024 exceeded fivefold compared to the preceding five-year period. While several review articles \cite{zhu2023heterophily,gong2024survey,luan2024heterophilic} have emerged during this timeframe, our work demonstrates distinctive value through the following dimensions:
\begin{itemize}
    \item In contrast to the prior efforts by Zhu et al. \cite{zhu2023heterophily} and Gong et al. \cite{gong2024survey} largely limited to pre-2023 methods, our survey presents three key advances: (1) systematic integration of cutting-edge developments through March 2025, (2) proposal of a more explicit taxonomy for heterophily GNNs with independent significance in categorization, and (3) complementary technical visualizations including schematic framework diagrams and inductive mathematical formulations.
    
    \item While the recent domain handbook by Luan et al. \cite{luan2024heterophilic} provides broad literature coverage, including many 2024 studies (primarily ArXiv preprints), our survey reveals two-fold differentiation:
    (1) classification clarity -- contrasting their simplified listing paradigm, we establish a clear hierarchical classification taxonomy; (2) temporal completeness - extending beyond their Q2-2024 cutoff, our survey incorporates critical March 2025 advancements.
\end{itemize}

In this paper, we present a comprehensive and systematic review of GNNs for heterophilic graphs, aiming to provide a general blueprint of heterophilic graph research. 
It can be beneficial to establish connections and make comparisons among different heterophilic GNN methods, leading to an in-depth understanding of how different methods tackle the challenges of heterophily learning.
We are expecting that our survey will significantly inspire and facilitate the development of heterophilic graphs\footnote{Our preprint of this article \cite{zheng2022graph} has attracted more than 420 citations, showing a strong uptrend of this research topic.}. The contributions of our work are summarized as follows: 
\begin{itemize}
    \item {\textbf{Comprehensive Overview}}: We provide a comprehensive overview of current heterophilic GNNs in terms of data, algorithms, and applications. We provide detailed descriptions of each model type, along with the necessary comparison and the gist summary.
    
    \item {\textbf{Systematic Taxonomy}}: We provide a systematic taxonomy of heterophilic GNNs and categorize existing methods into three classes, \ie, non-local neighbor extension methods, GNN architecture refinement methods, and hybrid methods. 

     \item {\textbf{Thorough Discussion}}: We provide a thorough discussion of the correlation between graph heterophily and various graph research domains, including the relation between graph heterophily and model robustness, over-smoothing, and graph anomaly detection.
    
    \item {\textbf{Future Directions}}: We suggest promising future research directions and discuss the limitations of existing heterophilic GNNs from multiple perspectives, namely interpretability, robustness, scalability, and heterophilic graph data exploration.
\end{itemize}

The remainder of this article is organized as follows. Section \ref{Preliminary} defines the related concepts and provides notations used in this survey. Section \ref{Framework} describes the framework of heterophilic GNNs and provides the taxonomy. Section \ref{Non-Local}-\ref{Hybrid} review three categories of heterophilic GNNs methods respectively. Section \ref{Discussion} discusses heterophily GNNs on diverse graphs and the correlation between heterophily and diverse graph research domains. Section \ref{Future} analyzes the unexplored challenges and potential future directions. Section \ref{Conclusion} concludes this article in the end. More details of real-world heterophilic graph dataset benchmarks, open-source codes, and the overall development timeline of heterophilic GNNs can be found in the appendix.

\section{Preliminary} \label{Preliminary}

\subsection{Notations}

Let $\mathcal{G} = (\mathcal{V}, \mathcal{E})$ be an undirected, unweighted graph where $\mathcal{V} = \{v_1, \cdots, v_{|\mathcal{V}|} \}$ is the node set and $\mathcal{E} \in \mathcal{V} \times \mathcal{V}$ is the edge set. The neighbor set of node $v$ is denoted as $\mathcal{N}(v) = \{u: (v, u) \in \mathcal{E}  \}$. The node features are represented by a feature matrix $\mathbf{X} \in \mathbb{R}^{|\mathcal{V}| \times d}$, where the $i$-th row $\mathbf{x}_i \in  \mathbb{R}^{d}$ is the feature vector of node $v_i$ and $d$ is the number of feature dimensions. Connectivity is represented by the adjacency matrix $\mathbf{A}\in \{0,1\}^{n\times n}$. For any matrix $\mathbf{A}$, we use $\mathbf{A}_{uv}$ to refer to the scalar value at the $(u,v)$ location. The graph Laplacian is defined as $\mathbf{L}=\mathbf{D}-\mathbf{A}$, where $\mathbf{D}\in \mathbb{R}^{n\times n}$ is the diagonal degree matrix.
Due to its generalization ability \cite{bollobas2004extremal}, the symmetric normalized Laplacian is often used, which is defined as $\Tilde{\mathbf{L}}=\mathbf{D}^{-1/2}\mathbf{L}\mathbf{D}^{-1/2}$. Another option is random walk normalization: $\Tilde{\mathbf{L}}=\mathbf{D}^{-1}\mathbf{L}$. Note that normalization could also be applied to the adjacency matrix. Their relationship can be derived as $\Tilde{\mathbf{L}}=\mathbf{I}-\Tilde{\mathbf{A}}$, where $\mathbf{I}$ is the identity matrix. Through eigendecomposition, $\mathbf{L}$ can be expressed as $\Tilde{\mathbf{L}}=\mathbf{U}\boldsymbol{\Lambda}\mathbf{U}^{\mathsf{T}}$, where each column of $\mathbf{U}\in \mathbb{R}^{n\times n}$ represents an eigenvector of $\mathbf{L}$, $\boldsymbol{\Lambda}$ is the diagonal matrix whose diagonal elements are the corresponding
eigenvalues (\ie, $\boldsymbol{\Lambda}_{ii}=\lambda_i$).


\subsection{Graph Neural Networks}
Generally, GNNs adopt the message passing mechanism, where each node representation is updated by aggregating the messages from local neighbor representations, and then combining the aggregated messages with its ego representation~\cite{gin_xu2019powerful}. The updating process of the $l$-th GNN layer for each node $v \in \mathcal{V}$ can be described as:

\begin{equation}
\label{eq:GNN}
\begin{aligned}
\mathbf{m}^{(l)}_v & = \operatorname{AGGREGATE}^{(l)}(\{\mathbf{h}^{(l-1)}_u: u \in \mathcal{N}(v)\}),
\\
\mathbf{h}^{(l)}_{v} & =\operatorname{UPDATE}^{(l)}(\mathbf{h}^{(l-1)}_v, \mathbf{m}^{(l)}_{v}),
\end{aligned}
\end{equation}

\noindent where $\mathbf{m}^{(l)}_v$ and $\mathbf{h}^{(l)}_{v}$ stand for the message vector and the representation vector of node $v$ at the $l$-th layer, respectively. And $\operatorname{AGGREGATE}(\cdot)$ and $\operatorname{UPDATE}(\cdot)$ are aggregation function (\eg, mean, LSTM, and max pooling) and update function (\eg, linear-layer combination)~\cite{sage_hamilton2017inductive}, respectively. Given the input of the first layer as $\mathbf{H}^{(0)} = \mathbf{X}$, the learned node representations at each layer of $L$-layer GNN can be denoted as $\mathbf{H}^{(l)} = \left[\mathbf{h}^{(l)}_{v}\right]$ for $v=(1,\cdots,\mathcal{|V|})$ and $ l=(1,\cdots,L)$. For node classification task, the final node representation $\mathbf{H}^{(L)}$ would be fed into a classifier network (\eg, a fully-connected layer) to generate the predictions for classes.
\subsection{Spectral Graph Convolution}
A graph convolution operation is defined in the Fourier domain such that
\begin{equation}
\label{eq:SpeGNN1}
f_1*f_2=\mathbf{U}[(\mathbf{U}^{\mathsf{T}}f_1)\odot (\mathbf{U}^{\mathsf{T}}f_2)],
\end{equation}
where $\odot$ is the element-wise product, and $f_1/f_2$ are two signals defined on nodes. It follows that a node signal $f_2=\mathbf{X}$ is filtered by spectral signal $\hat{f}_1=\mathbf{U}^{\mathsf{T}}f_1=\mathbf{g}$ as
\begin{equation}
\label{eq:SpeGNN2}
\mathbf{h}_v^{(l)}\!=\!\mathbf{g}(\Tilde{\mathbf{L}})\mathbf{h}_v^{(l-1)}\!=\!\mathbf{U}[\mathbf{g}(\boldsymbol{\Lambda})\odot (\mathbf{U}^{\mathsf{T}}\mathbf{h}_v^{(l-1)})]\!=\!\mathbf{U}\mathbf{g}(\boldsymbol{\Lambda})\mathbf{U}^{\mathsf{T}}\mathbf{h}_v^{(l-1)},
\end{equation}
where $\mathbf{g}$ is known as frequency response function. 
Therefore, the objective of spectral methods is to learn a function $\mathbf{g}(\cdot)$. In simpler terms, $\mathbf{g}(\cdot)$ can be seen as a way to re-weight signals of different frequencies (or eigenvalues). Eigenvalues represent the smoothness or frequency of the corresponding eigenvectors. Consequently, assigning greater weight to smaller eigenvalues retains more low-frequency information, while assigning greater weight to larger eigenvalues retains more high-frequency information. In general, $F_{LP}=\epsilon\mathbf{I}+\mathbf{D}^{-1/2}\mathbf{A}\mathbf{D}^{-1/2}=(\epsilon+1)\mathbf{I}-\mathbf{L}$ is a low-pass filter, while $F_{HP}=\epsilon\mathbf{I}-\mathbf{D}^{-1/2}\mathbf{A}\mathbf{D}^{-1/2}=(\epsilon-1)\mathbf{I}+\mathbf{L}$ denotes high-pass filter.

\subsection{Measure of Heterophily \& Homophily}
In general, heterophily and homophily of a graph $\mathcal{G} = (\mathcal{V}, \mathcal{E})$ can be measured by following metrics: node homophily~\cite{geom_pei2020geom}, edge homophily~\cite{h2gcn_zhu2020beyond}, class homophily~\cite{lim2021large}, adjusted homophily \cite{platonov2022characterizing}, and unbiased homophily \cite{mironovrevisiting}.

Node homophily and edge homophily are two essential measures. Concretely, the node homophily 
is the average proportion of the neighbors with the same class of each node:
\begin{equation}
\label{eq:node_homo}
\mathcal{H}_{node} = \frac{1}{|\mathcal{V}|} \sum_{v \in \mathcal{V}} \frac{|\{u \in \mathcal{N}(v): y_v = y_u \}|}{|\mathcal{N}(v)|}.
\end{equation}
The edge homophily 
is the proportion of edges connecting two nodes with the same class:
\begin{equation}
\label{eq:edge_homo}
\mathcal{H}_{edge} = \frac{|\{(v, u) \in \mathcal{E}: y_{v}=y_{u}\}|}{|\mathcal{E}|}.
\end{equation}
To alleviate the sensitivity issue of $\mathcal{H}_{node}$ and $\mathcal{H}_{edge}$ on the node class number, the class homophily measures the average proportion of neighbors with the same class across all nodes of each class while taking into account the constraint of class proportions, that is 
\begin{equation}
\label{eq:class_homo}
\mathcal{H}_{class}\!=\!\frac{1}{C-1}\sum\limits_{c=1}^C\left[\frac{\sum\limits_{v:y_v\!=\!c}|\{u \in \mathcal{N}(v)\!:\!y_v\!=\!y_u \}|}{\sum_{v:y_v=c}|\mathcal{N}(v)|}\!-\!\frac{n_c}{|\mathcal{V}|}\right]_{+},
\end{equation}
where $n_c=|\{u:y_u=c\}|$ denotes the node number of class $c$.

Platonov et al. \cite{platonov2022characterizing} and Mironov et al. \cite{mironovrevisiting} have progressively advanced homophily measurement by introducing metrics that fulfill an increasing number of desirable properties. Their first study \cite{platonov2022characterizing} proposed adjusted homophily, that is 
\begin{equation}
\label{eq:adjusted_homo}
\mathcal{H}_{adj} = \frac{\mathcal{H}_{edge}-\sum_{c=1}^C\left(\sum\limits_{v:y_v=c}|\mathcal{N}(v)|/(2|\mathcal{E}|)\right)^2}{1-\sum_{c=1}^C\left(\sum\limits_{v:y_v=c}|\mathcal{N}(v)|/(2|\mathcal{E}|)\right)^2},
\end{equation}
which is comparable across datasets with varying class statistics. 
And recent research \cite{mironovrevisiting} introduced unbiased homophily 
\begin{equation}
\label{eq:un}
\mathcal{H}_{un}=\frac{\sum_{i<j}(\sqrt{c_{ii}c_{jj}}-c_{ij})}{\sum_{i<j}(\sqrt{c_{ii}c_{jj}}+c_{ij})},
\end{equation}
where $c_{ii}=|\{(v, u) \in \mathcal{E}: y_{v}=y_{u}=i\}|/|\mathcal{E}|$ and $c_{ij}=|\{(v, u) \in \mathcal{E}: \{y_{v},y_{u}\}=\{i,j\}\}|/(2|\mathcal{E}|)$ for $i\neq j$. This metric is designed for reliable application across differing label distributions that satisfies the vast majority of these properties.

\noindent\textbf{Remark.} Node homophily $\mathcal{H}_{node}$ and edge homophily $\mathcal{H}_{edge}$ are still popular measurement methods among recent works.

\subsection{Real-World Benchmarks}\label{datasets}
Benchmark datasets are fundamental for evaluating model performance. The field was initially shaped by Pei et al. \cite{geom_pei2020geom}, who introduced six benchmark heterophilic graphs—Cornell, Texas, Wisconsin, Chameleon, Squirrel, and Actor—that remain the most widely adopted in current literature, despite their generally small scale. Building upon these, Lim et al. \cite{lim2021large,lim2021new} proposed a series of large-scale datasets, including ArXiv-Year, Snap-Patents, Penn94, Pokec, Genius, Deezer-Europe, Twitch-Gamers, and YelpChi. More recently, work in \cite{platonov2022critical} identified limitations such as small scale, class imbalance, and class leakage in earlier widely-used datasets, and subsequently introduced five high-quality alternatives: Roman-Empire, Amazon-Ratings, Minesweeper, Tolokers, and Questions. A detailed description of these datasets is provided in Appendix B.

\subsection{Mainstream Heterophily Learning Task}
The main objective learning task of heterophily GNNs is \textit{semi-supervised node classification}. 
In this task, we have a training graph $\mathcal{G} = (\mathbf{X}, \mathbf{A}, \mathbf{Y})$, where $\mathbf{X}\in \mathbb{R}^{N \times d}$ denotes $N$ nodes with $d$-dimensional features, and $\mathbf{A} \in \mathbb{R}^{N \times N}$ denotes the adjacency matrix indicating the edge connection. Assuming each node $v$ belongs to one out of $C$ classes, only a part of nodes are provided with labels as $\mathbf{Y}_v \in \mathcal{Y}_{L} = \{1,\cdots, C\}$ in the training node set. The goal of the task is to predict the classes of nodes whose labels are not given.

Concretely, given a heterophily GNN model parameterized by $\boldsymbol{\theta}$, denoting as $\text{GNN}_{\boldsymbol{\theta}}(\cdot)$. 
For semi-supervised node class
classification task, the cross-entropy loss is minimized to learn the optimal GNN parameters $\boldsymbol{\theta}$ over all labeled nodes as:
\begin{equation}
\begin{aligned}
\min_{\boldsymbol{\theta}}
\mathcal{L}_\text{cross-entropy}\left(\hat{\mathbf{Y}}, \mathbf{Y}\right),  \, \text{where} \, \, \hat{\mathbf{Y}}=\text{GNN}_{\boldsymbol{\theta}}(\mathbf{X},\mathbf{A}),
\end{aligned}
\end{equation}
where $\hat{\mathbf{Y}}=[\hat{\mathbf{Y}}_v]_{v\in\mathcal{V}}$ denotes GNN predicted node labels and $\mathbf{Y}=[\mathbf{Y}_v]_{v\in\mathcal{V}}$ is the ground-truth node labels.

\section{GNNs with Heterophily: Framework and Taxonomy} \label{Framework}

In this section, we provide a unified framework of heterophilic GNNs, and further categorize it from the lens of the message passing mechanism.

\subsection{Framework of heterophilic GNNs}

Following the general message passing principle for GNN model design, heterophilic GNNs focus on customizing the neighborhood aggregation and feature update schemes that specifically model the heterophily property. 
In contrast to homophilic GNNs, heterophilic GNNs exhibit distinct characteristics in three key design principles:

\textbf{P1: Non-locality of Neighbor Sets:} Incorporating information from non-local neighbors that may share the same class label as the central node;

\textbf{P2: Class Distinguishability:} Ensuring the aggregation process to effectively distinguish the class labels from both local and non-local neighbors;

\textbf{P3: Depth Fusion of Multi-layer Information:}  Integrating hierarchical messages from different inter layers of GNNs for capturing comprehensive heterophily property.


\subsection{Taxonomy of GNNs with Heterophily}
According to the above three-fold design principles in heterophily instructed neighbor aggregation and feature updating,
heterophilic GNNs can be categorized into three groups, including:

\textbf{(1) Non-local neighbor extension methods $\gets$ P1}.

\textbf{(2) GNN architecture refinement methods $\gets$ P2 \& P3}.

\textbf{(3) Hybrid methods $\gets$ P1, P2 \& P3}.

\begin{figure}[t!]
\centering
\includegraphics[width=0.498\textwidth]{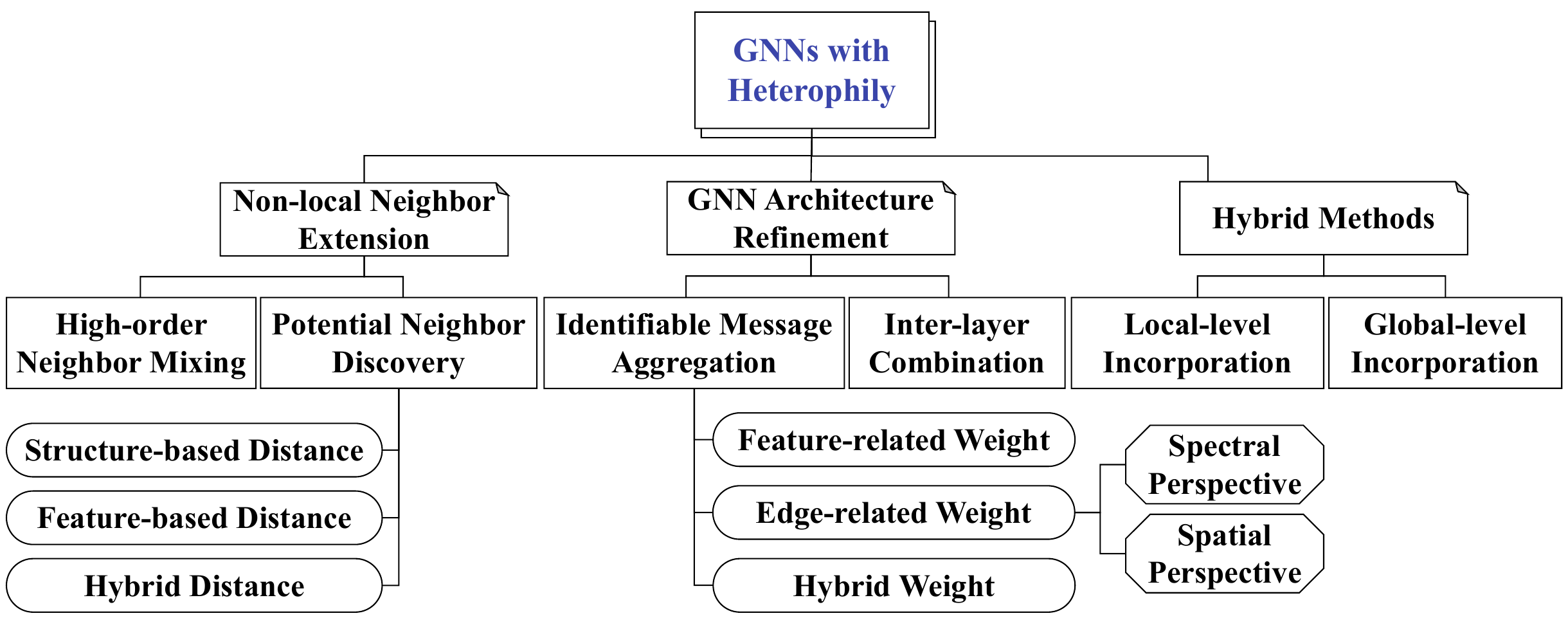}
\caption{Categorization of Heterophily GNNs.}\label{fig:Categorization}
\end{figure}

More fine-grained categorizations of these methods are briefly discussed below and shown in Fig.~\ref{fig:Categorization}. 


\noindent\textbf{Non-Local Neighbor Extension Methods} 
attempt to improve node representation by incorporating higher-order neighbor nodes that share labels during the message-passing process. This kind of methods break the locality limitations of $\mathcal{N}(v)$ in Eq. \eqref{eq:GNN} 
by reconstructing the non-local neighbor set from two perspectives: high-order neighbor mixing and potential neighbor discovery. Specifically, these methods focus on discovering appropriate neighbors from multi-hops nodes and redefining neighbor sets as $\mathcal{N}_p(v) = \{u: \operatorname{dist}(u, v) \leq p\}$ in the message aggregation stage in Eq. \eqref{eq:GNN}, where $\operatorname{dist}(u,v)$ denotes the distance method to measure the distance between nodes $u$ and $v$. $p$ is a threshold to limit the number of neighbors. In particular, $\mathcal{N}_p(v)$ can be degenerated into $\mathcal{N}(v)$ when the metric function $\operatorname{dist}(\cdot)$ is the shortest path distance and $p=1$.

\noindent\textbf{GNN Architecture Refinement Methods} boost the expressive ability of GNNs by facilitating message passing that distinguishes between similar and dissimilar neighbors or by deeply fusing multi-layer information. Specifically, focusing on the crafted architecture design of message aggregation and feature updates, these methods are categorized into identifiable message aggregation methods and inter-layer combination methods. Among these, identifiable message aggregation methods have gained significant attention. Their objective is to allocate appropriate weights in the message aggregation process, aiming to strengthen the fusion of homophilic information while mitigating the fusion of heterophilic information. With a focus on weight assignment schemes from various perspectives, existing methods can be further classified into three types: edge-related weight, feature-related weight, and hybrid weight.

\noindent\textbf{Hybrid Methods} can be taken as the combination of non-local neighbor extension methods and GNNs architecture refinement methods.
Typically, hybrid methods first construct an appropriate neighbor set through non-local neighbor extension, and then refine the GNN architectures from two perspectives: (1) local-level incorporation, that focuses on intra-layer heterophily-guided message passing, and (2) global-level incorporation, that enhances effective inter-layer heterophily information transfer throughout the entire GNN architecture. 

\noindent\textbf{Discussion.} Different heterophilic GNNs exhibit distinct properties. The fundamental concept of non-local neighbor extension methods is quite straightforward and simple to implement, as it directly models and captures the similarity relationships between nodes to enhance the homophily of local neighborhoods; however, for large-scale graphs, it can be memory-intensive to traverse the entire graph in order to model the similarity between each pair of nodes. GNN architecture refinement methods focus on mitigating the impact of local heterophilic information during the message passing process. This primarily involves further refinement of message aggregation and feature updates for the heterophilic design. 
However, the challenge lies in how to customize the heterophilic message passing mechanism with only a limited number of observable homophilic or heterophilic relationships.
Hybrid methods combine the advantages of both non-local neighbor extension and GNN architecture refinement. Nevertheless, a major challenge is devising a strategy for seamlessly integrating these approaches into the message-passing process.

\section{Heterophilic GNNs With Non-Local Neighbor Extension} \label{Non-Local}
\begin{figure}[t!]
\centering
\includegraphics[width=0.49\textwidth]{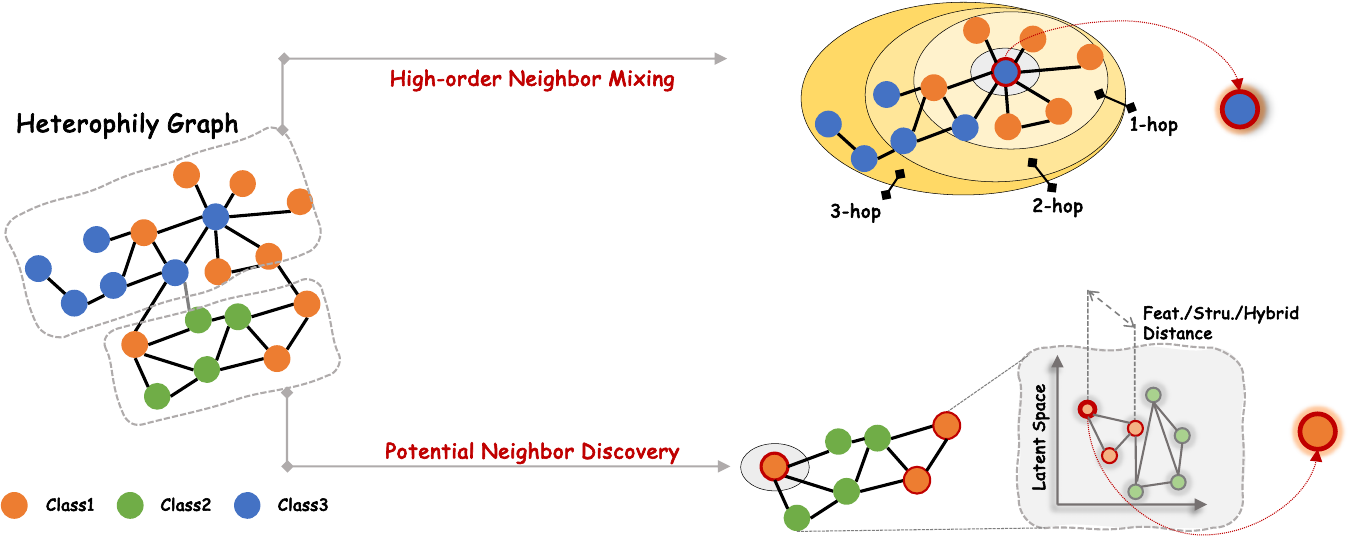}
\caption{Schematic diagram of high-order neighbor mixing method and potential neighbor discovery method.}\label{fig:HN_PN}
\end{figure}

Under the uniform message passing framework of homophilic GNNs, the neighborhood is usually defined as the set of all neighbors one-hop away (\eg, GCN), which means only messages from proximal nodes in a graph are aggregated. However, such a local neighborhood definition might not be appropriate for heterophilic graphs, where nodes belonging to the same class exhibit high structural similarity but can be farther away from each other. In light of these, current heterophilic GNNs attempt to extend the local neighbors to non-local ones primarily through two schemes: \textit{high-order neighbor mixing} and \textit{potential neighbor discovery}. As a result, the representation ability of heterophilic GNNs can be improved significantly by capturing the important features from distant and informative nodes. The pipelines of two example methods are given in Fig.~\ref{fig:HN_PN}, and a summary of the non-local neighbor extension works is illustrated in Table \ref{tab:main1}.

\subsection{High-order Neighbor Mixing}
Higher-order neighbor mixing allows the ego node to receive latent representations from their local one-hop neighbors and from further $k$-hop neighbors, so that the heterophilic GNNs can mix latent information from neighbors at various distances. Formally, the $k$-hop neighbor set is defined as
\begin{equation}
\label{eq:Non-Local_GNN1}
\mathcal{N}_k(v) = \{u: \operatorname{dist}(u, v)=k\}, 
\end{equation}
where $\operatorname{dist}(u,v)$ measures the shortest path distance between nodes $u$ and $v$. In addition, how to mix the information from the different $k$-hop neighbor sets is an important research point of higher-order neighbor mixing methods. 

Typically, MixHop \cite{abu2019mixhop} is a representative method that aggregates messages from multi-hop neighbors. Apart from one-hop neighbors, MixHop also considers two-hop neighbors for message propagation. After that, the messages acquired from different hops are encoded by different linear transformations and then mixed by concatenation. The $l$-th layer of MixHop for each node $v\in\mathcal{V}$ can be described as 
\begin{equation}
\label{eq:MixHop}
\begin{aligned}
\mathbf{m}^{(l)}_{v,k} & = \operatorname{AGGREGATE}^{(l)}(\{\mathbf{h}^{(l-1)}_u: u \in \mathcal{N}_k(v)\}), \\
\mathbf{h}^{(l)}_{v,k} & =\operatorname{UPDATE}^{(l)}(\mathbf{h}^{(l-1)}_v, \mathbf{m}^{(l)}_{v,k}), \\
\mathbf{h}^{(l)}_{v} & = \|_{k=2}\mathbf{h}^{(l)}_{v,k},
\end{aligned}
\end{equation}
where $\|_{k=2}$ means column-wise combination of information from $2$-hop neighbors. It can be noted that cross-level information fusion of MixHop is performed after the feature aggregation and updating among the same hop neighbors. 

Another perspective for cross-level information fusion is to merge cross-hop neighbors before the feature aggregation stage, and then aggregate and update multi-hop neighbors' features simultaneously. The most representative method is TDGNN \cite{wang2021tree}, which uses a tree decomposition method to disentangle neighborhood information in different layers and focus on adjusting the aggregation stage in Eq. \eqref{eq:GNN} to promote information fusion among different layers, modeled as 
\begin{equation}
\label{eq:TDGNN}
\mathbf{m}^{(l)}_{v} = \operatorname{AGGREGATE}^{(l)}(\{\mathbf{h}^{(l-1)}_u: u \in \bigcup\limits_{k=1}^K\mathcal{N}_k(v)\}), 
\end{equation}
where $K$ is the highest-hop setting the collection neighbor sets.

Recently, Ordered GNN \cite{song2022ordered} proposes to order the messages passing into the node representation, with specific blocks of neurons targeted for message passing within specific hops. This is achieved by aligning the hierarchy of the rooted-tree of a central node with the ordered neurons in its node representation. Based on simple design in the spectral domain, EvenNet \cite{lei2022evennet} discards messages from odd-order neighbors inspired by balance theory, deriving a graph filter with only even-order terms, which can be generalized to graphs of different homophily.

In summary, the high-order neighbor mixing methods straightforwardly include higher-order neighbors in local neighbor sets and devise an appropriate combination scheme to effectively integrate multi-order neighborhood information. Its objective is to alleviate the impact of local heterophily by incorporating richer homophilic information from higher-order neighborhoods.

\subsection{Potential Neighbor Discovery}
Compared to high-order neighbor mixing methods directly utilizing the inherent structural information from graphs, potential neighbor discovery methods reconsider the definition of neighbors in heterophilic graphs and build innovative structural neighbors through the entire topology exploration with heterophily. Apart from the original neighbor set, these methods construct a new potential neighbor set that can be
further formalized as 
\begin{equation}
\label{eq:Non-Local_GNN2}
\mathcal{N}_\rho(v) = \{u: \operatorname{dist}(v,u) < \rho\},
\end{equation}
where $\operatorname{dist}(u,v)$ is a metric function that measures the distance between nodes $u$ and $v$ in a specifically defined latent space and $\rho$ is a threshold to limit the number of neighbors. It is evident that distance measurement plays a pivotal role in identifying suitable potential neighbors. Current methods can be categorized into three main types based on the focal variables of distance measurement: structure-based distance, feature-based distance, and hybrid distance.

\subsubsection{Structure-based Distance}
Structure-based distance methods typically search for potential neighbors that meet measurement criteria within the geometric relationship latent space, which is defined by prior information about the graph topology. Typically, Geom-GCN~\cite{geom_pei2020geom} maps the input graph to a continuous latent space and defines the geometric relationships, \ie, split 2D Euclidean geometry locations, as the criteria to discover potential neighbors. Apart from inherent neighbors in original input graphs, neighbors that conform to the defined geometric relationships also participate in the message aggregation of GCN. Specifically, based on the graph and latent space, a structural neighborhood is built as $(\{\mathcal{N}(v),\mathcal{N}_{\rho}(v)\},\tau)$ underlying the relational operator $\tau$, where $\mathcal{N}(v)$ is the set of adjacent nodes of $v$ in the graph, $\mathcal{N}_{\rho}(v)$ is the set of potential neighbors from which the distance to $v$ is less than a pre-given parameter $\rho$ in the latent space, described as 
\begin{equation}
\label{eq:Geom-GCN}
\mathcal{N}_{\rho}(v) = \{u:\left\|\mathbf{h}_u-\mathbf{h}_v\right\|_2<\rho\},
\end{equation}
where $\mathbf{h}_u$ and $\mathbf{h}_v$ are the representations for nodes $u$ and $v$. $\|\cdot\|_2$ is the Euclidean norm to measure relative positions between two nodes. $\rho$ is determined from zero until the average cardinality of $\mathcal{N}_s(v)$ equals that of $\mathcal{N}(v)$, $\forall v \in \mathcal{V}$, that is, when the average neighborhood sizes in the graph and latent spaces are the same.


\subsubsection{Feature-based Distance}
Different from the structure-based distance methods that use structural information to determine potential neighbor nodes, feature-based distance methods determine connection relationships based on feature similarity. 
For instance, U-GCN~\cite{jin2021universal} and SimP-GCN~\cite{simp_jin2021node} choose top $k$ similar node pairs in terms of feature-level cosine similarity for each ego node to construct the neighbor set through the kNN algorithm. The potential neighbor set for this method can be built as
\begin{equation}
\label{eq:U/SimP-GCN}
\mathcal{N}_K(v) = \{u:\operatorname{TopK}(\operatorname{Cos}(\mathbf{h}_u,\mathbf{h}_v),K)\},
\end{equation}
where $\operatorname{Cos}(\cdot, \cdot)$ is usually defined the cosine similarity function, $K$ is the number of nearest neighbors set manually, $\operatorname{TopK}(\cdot)$ denotes a pooling operation for discovering potential neighbors with $K$ highest similarity. The recent work HES-GSL~\cite{wu2023homophily} still determines relationships of node pairs through cosine similarity, and further designs homophily-enhanced self-supervision to provide more supervision for similarity learning. 

\newcommand{\cmark}{\ding{51}}
\newcommand{\xmark}{\ding{55}}
\newcommand{\tabincell}[2]{\begin{tabular}{@{}#1@{}}#2\end{tabular}}

\begin{table*}[t]
	\centering
    \caption{Summary of non-local neighbor extension methods. \small{`Structure-based Distance', `Feature-based Distance', and `Hybrid Distancetance' take the structure-based scheme, feature-based scheme, and hybrid scheme as the distance metric for potential neighbor discovery, respectively.}}
	\resizebox{0.75\textwidth}{!}{
\begin{tabular}{lp{3cm}p{3cm}p{3cm}}
\toprule
\textbf{Publication Year and Venue} & \textbf{Method} & \textbf{Neighbor Range} & \textbf{Distance Measurement} \\\midrule
\multicolumn{4}{c}{High-Order Neighbors Mixing} \\\midrule
{[}2019 ICML{]}  & MixHop~\cite{abu2019mixhop}     & 2-Hops             & Shortest Path Distance      \\ 
{[}2021 CIKM{]}  & TDGNN~\cite{wang2021tree}               & Multi-Hops     & Shortest Path Distance      \\ 
{[}2022 NeurIPS{]} & EvenNet~\cite{lei2022evennet}  & Odd-Hops           & Shortest Path Distance    \\ 
{[}2023 ICLR{]}    & Ordered GNN \cite{song2022ordered}   & Multi-Hops      & Shortest Path Distance  \\
{[}2024 ICLR{]}    & MPformer \cite{anonymous2023mpformer}     & Multi-Hops        & Shortest Path Distance \\
{[}2024 ICDE{]}    & AMUD \cite{sun2024breaking}     & Multi-Hops        & Shortest Path Distance \\
{[}2025 ICASSP{]}    & AAGCN \cite{rey2025redesigning}     & Multi-Hops        & Shortest Path Distance \\
{[}2025 CIKM{]}    & PathLens \cite{goyal2025pathlens}     & Multi-Hops        & Shortest Path Distance
\\
\midrule
\multicolumn{4}{c}{Potential Neighbors Discovery}                            \\\midrule
{[}2020 ICLR{]} & Geom-GCN~\cite{geom_pei2020geom} & Multi-Hops   & Structure-based Distance            \\ 
{[}2021 WSDM{]} & SimP-GCN~\cite{simp_jin2021node} & Multi-Hops      & Feature-based Distance            \\ 
{[}2022 AAAI{]} & HOG-GNN~\cite{wang2021powerful}   & 2-Hops            & Hybrid Distance    \\ 
{[}2022 ICML{]} & GloGNN~\cite{li2022finding}       & Multi-Hops      & Hybrid Distance         \\ 
{[}2022 NeurIPS{]} & Diag-NSD~\cite{bodnar2022neural}     & Multi-Hops   & Feature-based Distance         \\ 
{[}2023 TNNLS{]} & HES-GSL~\cite{wu2023homophily}   & Multi-Hops       & Feature-based Distance        \\ 
{[}2023 WWW{]} & SE-GSL~\cite{zou2023se}   & Multi-Hops                   & Hybrid Distance         \\ 
{[}2023 ICML{]} & GOAL~\cite{zheng2023finding}    & Multi-Hops      & Hybrid Distance          \\ 
{[}2023 CVPR{]} & HopGNN~\cite{chen2023node}     & Multi-Hops            & Feature-based Distance        \\
{[}2024 IJCAI{]} & LG-GNN~\cite{yu2024lg} & Multi-Hops            & Hybrid Distance \\
{[}2025 CIKM{]} & DiRW~\cite{su2025dirw} & Multi-Hops            & Structure-based Distance
\\\bottomrule      
\end{tabular}
}
		\label{tab:main1}
\end{table*}

\begin{table*}[t]
	\centering
    \caption{Summary of GNN architecture refinement methods. \small{`Feature-Related', `Edge-Related' and `Hybrid' respectively represent weight assignment schemes that work on node feature, edges, and both of them for adaptive message aggregation. `\cmark' and `\xmark' indicate whether to include the according schemes, respectively.}}
	\resizebox{0.9\textwidth}{!}{
\begin{tabular}{lp{2.5cm}p{2.5cm}p{3.2cm}<{\centering}p{3.2cm}<{\centering}}
\toprule
\textbf{Publication Year and Venue} & \textbf{Method} & \textbf{Weight Assignment} & \textbf{Ego-Neighbor Separation} & \textbf{Combination Function}  \\\midrule
\multicolumn{5}{c}{Identifiable Message Aggregation} \\\midrule
{[}2021 AAAI{]} & FAGCN \cite{fagcn_bo2021beyond}                       & Edge-Related  & \xmark           & \xmark    \\
{[}2021 AAAI{]}    & CPGNN \cite{zhu2021graph}                          & Feature-Related      & \xmark        & \xmark    \\
{[}2021 NeurIPS{]} & DMP \cite{yang2021diverse}                         & Edge-Related  & \xmark          & \xmark    \\
{[}2021 ICDM{]}   & CGCN \cite{yan2021two}                              & Feature-Related      & \cmark        & \xmark    \\
{[}2022 ICLR{]}   & ACM \cite{acm_luan2021heterophily}                   & Edge-Related    & \xmark        & \xmark    \\
{[}2022 WWW{]}    & GBK-GNN \cite{du2022gbk}                            & Feature-Related      & \xmark        & \xmark    \\
{[}2022 WWW{]}    & F2GNN \cite{wei2022designing}                        & Edge-Related    & \xmark        & \xmark    \\
{[}2022 WWW{]}    & MWGNN \cite{ma2022meta}                             & Hybrid     & \xmark    & \xmark    \\
{[}2022 ICML{]}   & GIND \cite{chen2022optimization}                     & Feature-Related      & \cmark & \xmark    \\
{[}2022 NeurIPS{]}& ASGC \cite{chanpuriya2022simplified}                 & Edge-Related    & \xmark & \xmark    \\
{[}2022 LoG{]}    & GESN \cite{tortorella2022leave}                      & \xmark          & \cmark & \xmark    \\
{[}2022 TNNLS{]}  & NCGNN \cite{yang2022ncgnn}                           & Feature-Related      & \xmark & \xmark    \\
{[}2023 TNNLS{]}  & RFA-GNN \cite{wu2023beyond}                         & Edge-Related    & \xmark & \xmark    \\
{[}2023 TKDE{]}   & AutoGCN \cite{wu2023beyond2}                        & Edge-Related    & \xmark & \xmark    \\
{[}2023 ICML{]}   & GOAT \cite{kong2023goat}                            & Edge-Related    & \xmark & \xmark    \\
{[}2023 WWW{]}    & Mid-GCN \cite{huang2023robust}                       & Edge-Related    & \xmark & \xmark    \\
{[}2023 NeurIPS{]}& TEDGCN \cite{yan2023from}                           & Edge-Related    & \xmark & \xmark    \\
{[}2023 NeurIPS{]}& LRGNN \cite{liang2023predicting}                     & Edge-Related    & \xmark & \xmark    \\ 
{[}2024 AAAI{]}& AHGFC \cite{wen2024homophily}                     & Edge-Related    & \xmark & \xmark    \\ 
{[}2024 AAAI{]}& PC-Conv \cite{li2024pc}                     & Edge-Related    & \xmark & \xmark    \\ 
{[}2024 WWW{]}& VR-GNN \cite{shi2024vr}                     & Edge-Related    & \xmark & \xmark    \\
{[}2024 NeurIPS{]}& TFE-GNN \cite{duan2024unifying}                     & Edge-Related    & \xmark & \xmark    \\ 
{[}2024 TNNLS{]}& PEGFAN \cite{li2024permutation}                     & Edge-Related    & \xmark & \xmark    \\ 
{[}2024 TKDE{]}& DHGR \cite{bi2024make}                     & Edge-Related    & \xmark & \xmark    \\
{[}2024 TCYB{]}& Flow2GNN \cite{huang2024flow2gnn}                     & Hybrid    & \xmark & \xmark    \\
{[}2025 ICASSP{]} & RETAIN \cite{zou2025retain}                     & Edge-Related    & \xmark & \xmark    \\
{[}2025 IJCAI{]} & H$_2$OGNN \cite{wang2025all}                     & Edge-Related    & \xmark & \xmark    \\
{[}2025 ACM TIS{]} & GSF-GNN \cite{liu2025global}                     & Hybrid    & \xmark & \xmark    \\
{[}2025 KDD{]} & RASH \cite{zheng2025enhancing}                     & Edge-Related    & \xmark & \xmark    \\\midrule
\multicolumn{5}{c}{Inter-Layer Combination}           \\\midrule
{[}2018 ICML{]}    & JK-Net \cite{xu2018representation}                  & \xmark        & \xmark           & Skip Connections           \\
{[}2020 ICLR{]}    & GPR-GNN \cite{chien2020adaptive}                    & \xmark        & \xmark           & Adaptive Connections        \\
{[}2022 NeurIPS{]} & PowerEmbed \cite{huang2022local}                   & \xmark        & \xmark           & Skip Connections  
\\\bottomrule
\end{tabular}}
		\label{tab:main2}
\end{table*}

\subsubsection{Hybrid Distance}
Recent works increasingly emphasize the comprehensive consideration of structural and node feature information in potential neighbor discovery. These methods focus more on the potential message passing among global homophily neighbors in the graph. Typically, HOG-GCN \cite{wang2021powerful} and GloGNN \cite{li2022finding} design similarity calculation modules to capture the correlations between global nodes through both topological and attribute information. Specifically, HOG-GCN constructs a homophily degree matrix with the label propagation technique to explore the extent to which a pair of nodes belong to the same class in the entire heterophilic graph from the perspective of topology space. By involving the class-aware information during the propagation process, intra-class nodes with higher heterophily (\ie, lower homophily degree) would contribute more to the neighbor aggregation than underlying inter-class nodes. More simplified than HOG-GCN, GloGNN directly measures potential correlations between global nodes in terms of both feature attribute similarity and topology similarity by further regularizing node representations with nodes' multi-hop reachabilities. In general, the potential neighbors set can be defined as
\begin{equation}
\label{eq:HOG-GCN/GloGNN}
\mathcal{N}(v) = 
\mathcal{N}_{T}(v)\cap\mathcal{N}_{S}(v),
\end{equation}
where $\mathcal{N}_{T}(v)=\{u: \Tilde{\mathbf{h}}_u\Tilde{\mathbf{h}}_v^{\mathsf{T}}\neq 0\}$ and $\mathcal{N}_{S}(v)=\{u: \hat{\mathbf{h}}_u\hat{\mathbf{h}}_v^{\mathsf{T}}<\rho\}$ are potential neighbor sets defined by measuring node correlations in terms of topology similarity and feature similarity, $\cap$ is the intersection operation, and  $\Tilde{\mathbf{h}}_u$ and $\hat{\mathbf{h}}_u$ are representations of node $u$ through topology structure information (\ie, $\mathbf{A}$) and node attribute information (\ie, $\mathbf{X}$), respectively.

Furthermore, SE-GSL \cite{zou2023se} offers an effective measure of the information embedded in an arbitrary graph and structural diversity (where $\mathcal{N}_{T}(v)$ is characterized by both the kNN structure and the original topology structure) and presents a novel sampling-based mechanism for restoring the graph structure via node structural entropy distribution. It increases the connectivity among nodes with larger uncertainty in lower-level communities. GraphTU \cite{gao2023topology} offers a probabilistic approach to exploring potential neighbors. A key observation in this research is that the statistical variances based features and topology information within local neighborhoods can be effectively harnessed to extend the training distribution, creating novel potential neighbor sets through a non-parametric method. This approach is particularly valuable for addressing heterophily within the entire graph, especially in the case of minor class nodes. To reveal attribute relationships among nodes in the entire graph, GOAL \cite{zheng2023finding} augments the existing graph by constructing a fully connected graph through a graph completion process. This augmentation considers two modes: homophilic connections and heterophilic connections. An important aspect involves developing a method to distinguish between connections that reflect a tendency for homophily and those that indicate heterophily. This distinction is made based on the Connected Structure Difference (CSD) computed between connected node pairs and randomly selected node pairs, ultimately facilitating the optimization of the complemented graph.

In summary, potential neighbor discovery methods focus on identifying homophilic neighbors among high-order neighbors and incorporating them into local neighbor sets to enhance homophilic message-passing. 
By measuring the similarity of node features with different distance calculation schemes, potential neighbor discovery methods are able to effectively identify potential homophilic neighbors that share the same class labels.





\section{Heterophilic GNN Architecture Refinement} \label{Architecture-Refinement}
General GNN architectures in Eq.~(\ref{eq:GNN}) contain two essential components: the aggregation function $\operatorname{AGGREGATE}(\cdot)$ to integrate information from the discovered neighbors, and the update function $\operatorname{UPDATE}(\cdot)$ to combine the learned neighbor messages with the initial ego representation. Given the original local neighbors and the extended non-local neighbors on heterophilic graphs, existing GNN architecture refinement methods contribute to fully exploiting the neighbor information from the following aspects by accordingly revising $\operatorname{AGGREGATE}(\cdot)$ and $\operatorname{UPDATE}(\cdot)$: (1) {\it{Identifiable message aggregation}} discriminates and enhances the messages of similar neighbors from dissimilar ones; (2) {\it{Inter-layer combination}} emphasizes the effect of different propagation ranges (\ie, the number of GNN layers) on node representation learning. All these two aspects come to the same destination: boosting the expressive ability of GNNs for heterophilic graphs by encouraging distinguishable and discriminative node representations. 

\subsection{Identifiable Message Aggregation}
\begin{figure*}[!t]
\centering
{\includegraphics[width=\textwidth]{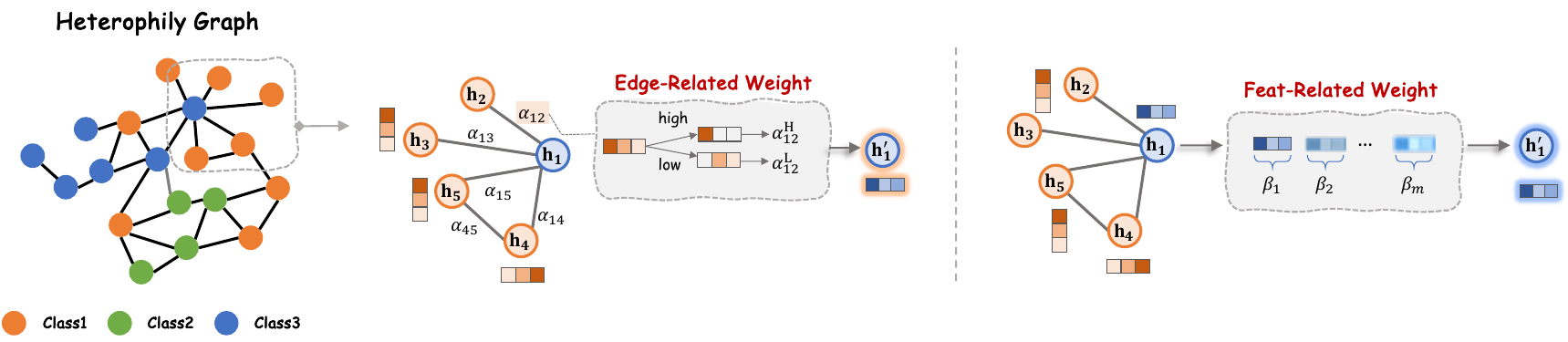}}
\caption{Illustration of identifiable message passing with edge-related weight and feature-related weight assignment schemes in GNN architecture refinement methods.}\label{fig:ERW_FRW}
\end{figure*}

Given the neighbors to be aggregated, the key of integrating beneficial messages on heterophilic graphs is distinguishing the information of similar neighbors (likely in the same class) from that of dissimilar neighbors (likely in different classes). To make node representations on heterophilic graphs more discriminative, identifiable message aggregation methods alter the aggregation operation $\operatorname{AGGREGATE}(\cdot)$ by imposing adaptive edge-aware weights $a_{uv}^{(l)}$ for node pair $(u,v)$ at the $l$-th layer as:
\begin{equation}
\label{eq:new_aggr}
\mathbf{m}^{(l)}_v  = \operatorname{AGGREGATE}^{(l)}(\{a_{uv}^{(l)}\,\mathbf{h}^{(l-1)}_u: u \in \mathcal{N}(v)\}).
\end{equation}
In this way, different methods develop various weight assignment schemes for $a_{uv}^{(l)}$ to model the importance of similar and dissimilar neighbors during aggregation. In the following, we provide the details of weight assignment schemes adopted by existing methods, which respectively work on node \textit{feature} and \textit{edge} related weights. Fig. \ref{fig:ERW_FRW} provides the pipelines of two weight assignment schemes.


\subsubsection{Edge-Related Weight}
In general, edge-related methods simultaneously focus on \textit{spectral} domain and the \textit{spatial} domain. To be concrete, spectral GNNs leverage the theory of graph signal processing to design graph filters, and spatial GNNs focus on the graph structural topology to develop aggregation strategies. 

\noindent\textbf{Spectral Perspective. }In contrast to Laplacian smoothing~\cite{li2018deeper} and low-pass filtering~\cite{sgc_wu2019simplifying} to approximate graph Fourier transformation on homophilic graphs, spectral GNNs on heterophilic graphs involve both low-pass and high-pass filters to adaptively extract low-frequency and high-frequency graph signals. The essential intuition behind this lies in that low-pass filters mainly retain the commonality of node features, while high-pass filters capture the difference between nodes. 

Typically, FAGCN~\cite{fagcn_bo2021beyond} adopts a self-gating attention mechanism to learn the proportion of low-frequency and high-frequency signals by splitting the $a_{uv}^{(l)}$ into two components, \ie, $a_{uv}^{(l,LP)}$ and $a_{uv}^{(l,HP)}$, corresponding to low-pass and high-pass filters, respectively. Through adaptive frequency signal learning, FAGCN could achieve expressive performance on different types of graphs with homophily and heterophily. Formally, adaptive edge-aware weights $a_{uv}^{(l)}$ are defined in FAGCN as
\begin{equation}
\label{eq:FAGCN}
a_{uv}^{(l)}=\frac{a_{uv}^{(l,LP)}-a_{uv}^{(l,HP)}}{\sqrt{d_ud_v}},
\end{equation}
where $d_u$ and $d_v$ respectively denote the degree of node $u$ and $v$, $a_{uv}^{(l,LP)}-a_{uv}^{(l,HP)}$ is the coefficient learned through a shared self-gating mechanism $\tanh{(\mathbf{q}^{\mathsf{T}}[\mathbf{h}_u;\mathbf{h}_v])}$, from which $[;]$ denotes the concatenation operation, $\mathbf{g}$ can be seen as a shared convolutional kernel, $\tanh{(\cdot)}$ is the hyperbolic tangent function, which can naturally limits the value of $a_{uv}^{(l,LP)}-a_{uv}^{(l,HP)}$ in $[-1, 1]$. In a similar fashion, RFA-GNN \cite{wu2023beyond} captures edge-aware weights using a relation-based frequency adaptive mechanism. This mechanism takes into account higher-order contextual information compared to FAGCN. It precisely defines edge-aware weights for a specific relation. Specifically, the edge-aware weight of $(l+1)$-order information between nodes $u$ and $v$ for relation $k$ is described as $a_{uvk}^{(l)}=\tanh{(\mathbf{q}_{lk}^{\mathsf{T}}[\mathbf{h}_{uk}^{(l)};\mathbf{h}_{vk}^{(l)}])}$, where $\mathbf{q}_{lk}$ is the attention coefficient for relation $k$ in the $l$-th iteration.

Apart from low-pass and high-pass filters, ACM~\cite{acm_luan2021heterophily} further involves the identity filter, which is the linear combination of low-pass and high-pass filters, i.e., 
\begin{equation}
\label{eq:ACM}
a_{uv}^{(l)}=\operatorname{Softmax}\left(([a_{uv}^{(l,LP)},a_{uv}^{(l,IP)},a_{uv}^{(l,HP)}]/\tau)\mathbf{W}_{uv}^{(l)}\right),
\end{equation}
where $\tau$ indicates a temperature parameter and $\mathbf{W}_{uv}^{(l)} \in\mathbb{R}^{1\times 3}$ is used to learn which filters is important or not for each node. $a_{uv}^{(l,LP)}$, $a_{uv}^{(l,IP)}$ and $a_{uv}^{(l,HP)}$ respectively denotes low-pass, identity and high-pass edge-aware weights, which learn from 3 channels features. AutoGCN \cite{wu2023beyond2} captures the full spectrum of graph signals and automatically update the bandwidth of graph convolutional filters. In addition, Mid-GCN~\cite{huang2023robust} contains a mid-pass filter determined by both low-pass and high-pass filters. The robustness of signals passing through this mid-pass filter is theoretically guaranteed by their analyses. 

In this way, ACM, AutoGCN and Mid-GCN could adaptively exploit beneficial neighbor information from different filter channels for each node; Meanwhile, their identity or mid-pass filters could guarantee less information loss of the input signal.

\noindent\textbf{Spatial Perspective. }Heterophilic GNNs in the spatial domain require the diverse topology-based aggregation of neighbors from the same or different classes guided by the heterophily.
Therefore, the edge-aware weights of neighbors should be assigned according to the spatial graph topology and node labels. Taking node attributes as weak labels, DMP method~\cite{yang2021diverse} considers node attribute heterophily for diverse message passing and specifies every attribute propagation weight on each edge.
Furthermore, instead of the scalar weight $a_{uv}^{(l)}$ that aggregates all the node attributes with the same weight at the node level, DMP extends the weight to a vector ${\bf{a}}_{uv}^{(l)}$ through operating in the attribute dimension, and this vector can be calculated by either relaxing GAT~\cite{velivckovic2018graph} weights to real values or allowing an element-wise average of neighbor weights.

Moreover, certain methods learn edge-aware weights based on node labels. For instance, Chen et al. \cite{chen2022graph} introduce the concept of graph decoupling attention Markov networks (GDAMNs). This approach incorporates variational inference to model edge uncertainty and employs both hard and soft attention mechanisms on node labels to improve the learning of edge-aware weights, represented as
\begin{equation}
\label{eq:GDAMNs}
a_{uv}^{(l)}=\frac{a_{uv}^{\mathsf{hard}}\exp{\delta_{uv}^{(l-1)}}}{\sum_{v=1}^Na_{uv}^{\mathsf{hard}}\exp{\delta_{uv}^{(l-1)}}},
\end{equation}
where $a_{uv}^{\mathsf{hard}}=\mathbf{A}_{uv}\cdot\Tilde{\mathbf{y}}^\mathsf{T}w_{uv}\Tilde{\mathbf{y}}$ is the hard attention with label similarity and $\delta_{uv}^{(l-1)} = -\operatorname{Cos}(\mathbf{h}_u^{(l-1)},\mathbf{h}_v^{(l-1)})$ denotes the soft attention with feature dissimilarity. In general, the hard attention is learned on labels for a refined graph structure with fewer inter-class edges so that the aggregation’s negative disturbance can be reduced. The soft attention aims to learn the aggregation weights based on features over the refined graph structure to enhance information gains during message passing.

Recent research has shown a growing interest in the efficiency of computing edge-aware weights for large-scale graphs. CGP \cite{liu2023comprehensive} introduces a graph pruning paradigm during training, enabling the discovery of high-performing sparse GNNs within a single training process. Additionally, a global graph transformer model has been proposed for large-scale heterophilic node classification tasks, known as GOAT \cite{kong2023goat}. GOAT samples potential neighbors for each node from its k-hop neighbors. To distinguish between neighbors based on topology information during the propagation process, GOAT further incorporates a global attention module designed to learn attention scores between potential neighbors and source nodes.

\subsubsection{Feature-Related Weight}
Apart from the adaptive edge-aware weight learning, there is another solution working on the neighbor representation learning by revising $\mathbf{h}^{(l-1)}_u$ in $\operatorname{AGGREGATE}^{(l)}(\{a_{uv}^{(l)}\,\mathbf{h}^{(l-1)}_u: u \in \mathcal{N}(v)\})$.
Conventionally, the above-mentioned methods mainly utilize the contextual node representations of neighbors; In contrast, this solution transforms contextual node embeddings $\mathbf{h}^{(l-1)}_u$ to other node-level properties that reflect the heterophily. In this way, heterophilic GNNs learn node-level attention $\beta_{v}^{(l)}$, which is shared with each $u\in \mathcal{N}(v)$, to capture the beneficial information of heterophily for distinguishable node representation learning. As a result, the aggregation progress can be redefined as $\operatorname{AGGREGATE}^{(l)}(\{\beta_{v}^{(l)}\,\mathbf{h}^{(l-1)}_u: u \in \mathcal{N}(v)\})$. Existing methods mainly learn $\beta_{v}^{(l)}$ through two schemes based on homophily/heterophily prior knowledge and homophily attributes attention.

\noindent\textbf{Prior-Based.}
The prior-based method considers the incorporation of homophily/heterophily prior knowledge in the process of feature transformation, primarily estimating the assignment weight $\beta_{v}^{(l)}$ through class information. Typically, instead of propagating original node feature representations, CPGNN~\cite{zhu2021graph} propagates a prior belief estimation based on a compatibility matrix, so that it can capture both heterophily and homophily by modeling the likelihood of connections between nodes in different classes. The feature-aware weight of embeddings $\mathbf{h}^{(l-1)}_v$ with prior belief is defined as
\begin{equation}
\label{eq:CPGNN}
\beta_v^{(l)}=\operatorname{Sinkhorn-Knopp}(y_v,\mathbf{A}_v, \mathbf{B}_v),
\end{equation}
where $\operatorname{Sinkhorn-Knopp}$ denotes the Sinkhorn-Knopp algorithm proposed in \cite{sinkhorn1967concerning}, which is
to ensure that $\beta_v^{(l)}$ is doubly stochastic. $\mathbf{B}_v$ denotes an enhanced belief matrix for node $v$, depending on a training mask matrix and a prior belief. 
Going beyond the uniform GCN aggregation, NLGNN~\cite{Liu2021NLGNN} and GPNN~\cite{yang2021graph} consider the sequential aggregation where the neighbors are ranked based on the class similarity (\ie, a homophily/heterophily prior belief) in order. A regular 1D-convolution layer is applied to extract the affinities between the sequential nodes whether the nodes are close or distant in heterophilic graphs. GIND \cite{chen2022optimization} extends the linear isotropic diffusion to a more expressive nonlinear diffusion mechanism, which learns nonlinear flux features between node pairs before aggregation. This design of the nonlinear diffusion ensures that more information can be aggregated from similar neighbors and less from dissimilar neighbors by learning $\beta_{v}^{(l)}$, making node features less likely to over-smooth. 

\noindent\textbf{Attention-Based.}
The majority of research focuses on developing diverse attention mechanisms for learning $\beta_{v}^{(l)}$. A straightforward approach involves learning attribute similarity and utilizing this similarity to determine attention on node features. CGCN~\cite{yan2021two} uses the cosine similarity to send signed neighbor features under certain constraints of relative node degrees. In this way, the messages are allowed to be optionally multiplied by a negative sign or a positive sign, \ie, for node $v$, its message is described as 
\begin{equation}
\begin{aligned}
\label{eq:GGCN}
\mathbf{h}^{(l+1)}_v = \sigma\left(\beta_0^{(l)}\Tilde{\mathbf{h}}^{(l)}_v+\beta_1^{(l)}(\mathbf{s}^{(l)}_\mathsf{v,pos}\odot \mathbf{A}_v^{(l)})\Tilde{\mathbf{h}}^{(l)}_v\right. \\
\left.+\beta_2^{(l)}(\mathbf{s}^{(l)}_\mathsf{v,neg}\odot \mathbf{A}_v^{(l)})\Tilde{\mathbf{h}}^{(l)}_v\right),
\end{aligned}
\end{equation}
where $\beta_0^{(l)}$, $\beta_1^{(l)}$, and $\beta_2^{(l)}$ are the $l$-th layer learned scalars. $\mathbf{s}^{(l)}_\mathsf{u,pos}$ and $\mathbf{s}^{(l)}_\mathsf{u,neg}$ respectively indicate positive matrix and negative matrix, which are split from the matrix with sign information. Intuitively, signed messages consist of the negated messages sent by neighbors of the opposing classes, and the positive messages sent by neighbors of the same class. Similarly, Du et al. \cite{du2022gbk} propose a method for defining positive and negative correlation weights when modeling the similarity and dissimilarity between node features. They introduce a novel GNN model called GBK-GNN, which is based on a bi-kernel feature transformation and a selection gate. The two kernels capture homophily and heterophily information, and the gate is used to determine which kernel should be applied to a specific pair of nodes. 
Formally, the gate signal is described as $\beta_v^{(l)}=\sum\limits_{u\in \mathcal{N}(v)}\operatorname{Sigmoid}\left(\operatorname{MLP}(\mathbf{h}_v^{(l-1)},\mathbf{h}_u^{(l-1)};\mathbf{W}_v^{(l-1)})\right)$, where $\operatorname{MLP}(\cdot)$ is a multilayer perceptron.
MMP \cite{chen2022memory} decouples the messages into two parts, i.e., memory for propagation and self-embedding for discrimination. The memory for propagation aims to endow each node with a memory cell and sends messages from the memory cell instead of hidden self-embedding. After propagation, each node can leverage a learnable control mechanism to adaptively update its self-embedding and memory cell according to their recent states. CAGNNs \cite{chenexploiting} investigates the feature aggregation of inter-class edges from an entire neighbor identifiable perspective by a new metric based on von Neumann entropy. Using an importance score $\beta_{v}^{(l)}$ obtained by a mixer combines discriminant feature and neighbor information, which are decoupled from embedding $\mathbf{h}^{(l-1)}_v$. To overcome the limitations of traditional messaging methods, NCGNN \cite{yang2022ncgnn} represents nodes as collections of node-level capsules. Each capsule is responsible for extracting distinct features from its associated node. For each node-level capsule, a novel dynamic routing procedure is implemented to intelligently select the most suitable capsules for aggregation from a subgraph identified by the designed graph filter. NCGNN exclusively aggregates advantageous capsules while constraining irrelevant messages to prevent excessive feature mixing among interacting nodes. As a result, this approach mitigates the problem of oversmoothing and enables the learning of effective node representations in graphs characterized by either homophily or heterophily.

\begin{figure*}[t!]
\centering
{\includegraphics[width=\textwidth]{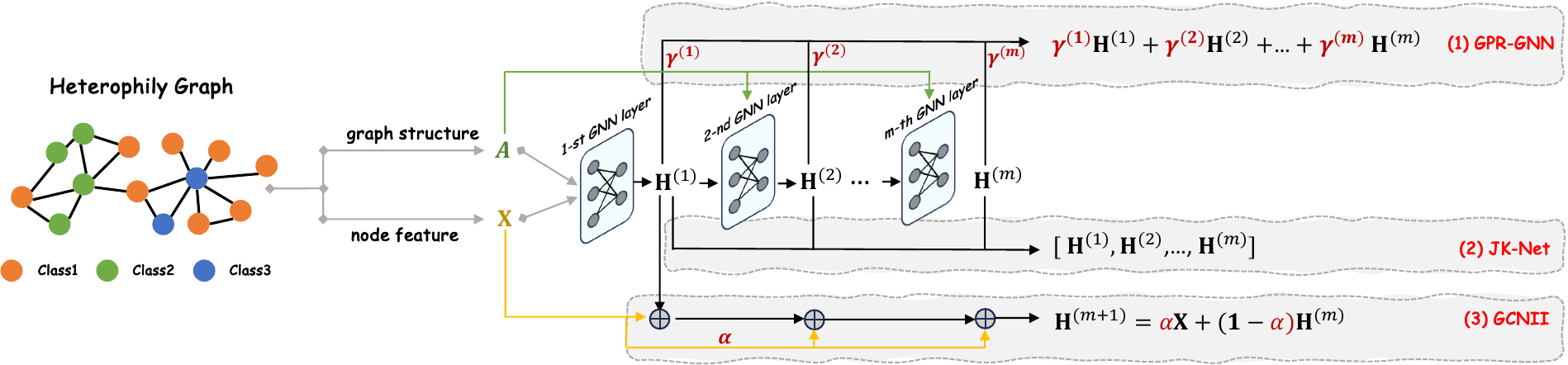}}
\caption{Illustration of typical inter-layer combination methods: (1) GPR-GNN~\cite{chien2020adaptive}; (2) JK-Net~\cite{xu2018representation}; and (3) GCNII~\cite{chen2020simple}.}\label{fig:IC}
\end{figure*}

\subsubsection{Hybrid Weight} Hybrid methods aim to calculate edge weights guided by node features and edge information simultaneously. A typical method is MWGNN \cite{ma2022meta}, which models the node local distribution from node feature, topological structure, and positional identity aspects with the meta-weight. 
Then, based on the meta-weight, an adaptive graph convolution is derived to conduct node-specific weighted aggregation for boosted node representation learning.

\noindent\textbf{Remark.}
On heterophilic graphs, an ego node is likely to be dissimilar with its neighbors in terms of class labels. Hence, encoding ego-node representations separately from the aggregated representations of neighbor nodes would benefit distinguishable node embedding learning. In detail, ego-neighbor separation methods detach self-loop connections of the ego nodes in $\operatorname{AGGREGATE}(\cdot)$. Meanwhile, they alter the $\operatorname{UPDATE}(\cdot)$ as the non-mixing operations, \eg, concatenation, instead of the mixing operation, \eg, ``average'' in vanilla GCN. H2GCN~\cite{h2gcn_zhu2020beyond} first proposes to exclude the self-loop connection and points out that the non-mixing operations in the update function ensure that expressive node representations would survive over multiple rounds of propagation without becoming prohibitively similar. Besides, WRGNN~\cite{SureshBNLM21} imposes different mapping functions on the ego-node embedding and its neighbor aggregated messages, while GGCN~\cite{yan2021two} simplifies the mapping functions to learnable scalar parameters to separately learn ego-neighbor representations. Moreover, ACM~\cite{acm_luan2021heterophily} adopts the identity filter to separate the ego embedding and then conducts the channel-level combination with its neighbor information in the update function.

\subsection{Inter-Layer Combination} 
Different from identifiable message aggregation methods focusing on the fine-grained intra-layer design of GNN architectures with heterophily, inter-layer combination methods consider layer-wise operations to boost the representation power of heterophilic GNNs. 
The intuition behind this strategy is that in the shallow layers of GNNs, they collect local information, \eg, one-hop neighbor locality in two-layer vanilla GCN, when the layers go deeper, GNNs gradually capture global information implicitly via multiple rounds of neighbor propagation. Due to the heterophily characteristic,
neighbors with similar information, \ie, class labels, might locate in both local geometry and long-term global topology. Hence, combining intermediate representations from each layer contributes to leveraging different neighbor ranges with the consideration of both local and global structural properties, resulting in powerful heterophilic GNNs. Fig. \ref{fig:IC} shows the formulation and illustrations of the closely related methods.

The prior idea first comes from JK-Net~\cite{xu2018representation} which flexibly captures better structure-aware representation with different neighborhood ranges.
The combination of features among different layers is described as
\begin{equation}
\label{eq:JKNet}
\hat{\mathbf{h}}_v = \operatorname{LA}(\mathbf{h}^{(1)},\mathbf{h}^{(2)},\ldots,\mathbf{h}^{(L)}),
\end{equation}
where $\operatorname{LA}(\cdot)$ denotes the layer aggregation, such as column-wise combination, max pooling, attention with LSTM, etc.

Compared with the methods using all previous intermediate representations, GCNII~\cite{chen2020simple} only integrates the first layer's node embedding at each layer with the initial residual connection, defined as
\begin{equation}
\label{eq:GCNII}
\mathbf{h}_v^{(l)}\!=\!\alpha\mathbf{h}_v^{(0)}+\operatorname{AGGREGATE}^{(l)}(\{(1-\alpha)\mathbf{h}^{(l-1)}_v\!:\!u\!\in\!\mathcal{N}(v)\}),
\end{equation}
where $\alpha$ is a hyperparameter that maintains a balance between the node embedding of the first layer and the representation of the current layer.

Instead of using the simple concatenation operation, GPR-GNN~\cite{chien2020adaptive} further assigns learnable weights to combine the representations of each layer adaptively via the Generalized PageRank (GPR) technique. PowerEmbed~\cite{huang2022local} employs an inception network to learn the rich representations that interpolate from local message-passing features to global spectral information. Hence, inter-layer combination methods are able to conduct topological feature exploration and benefit from informative multi-round propagation, making node features of heterophilic graphs distinguishable.


\section{Hybrid Methods}
\label{Hybrid}
\begin{figure}[t!]
\centering
\includegraphics[width=0.49\textwidth]{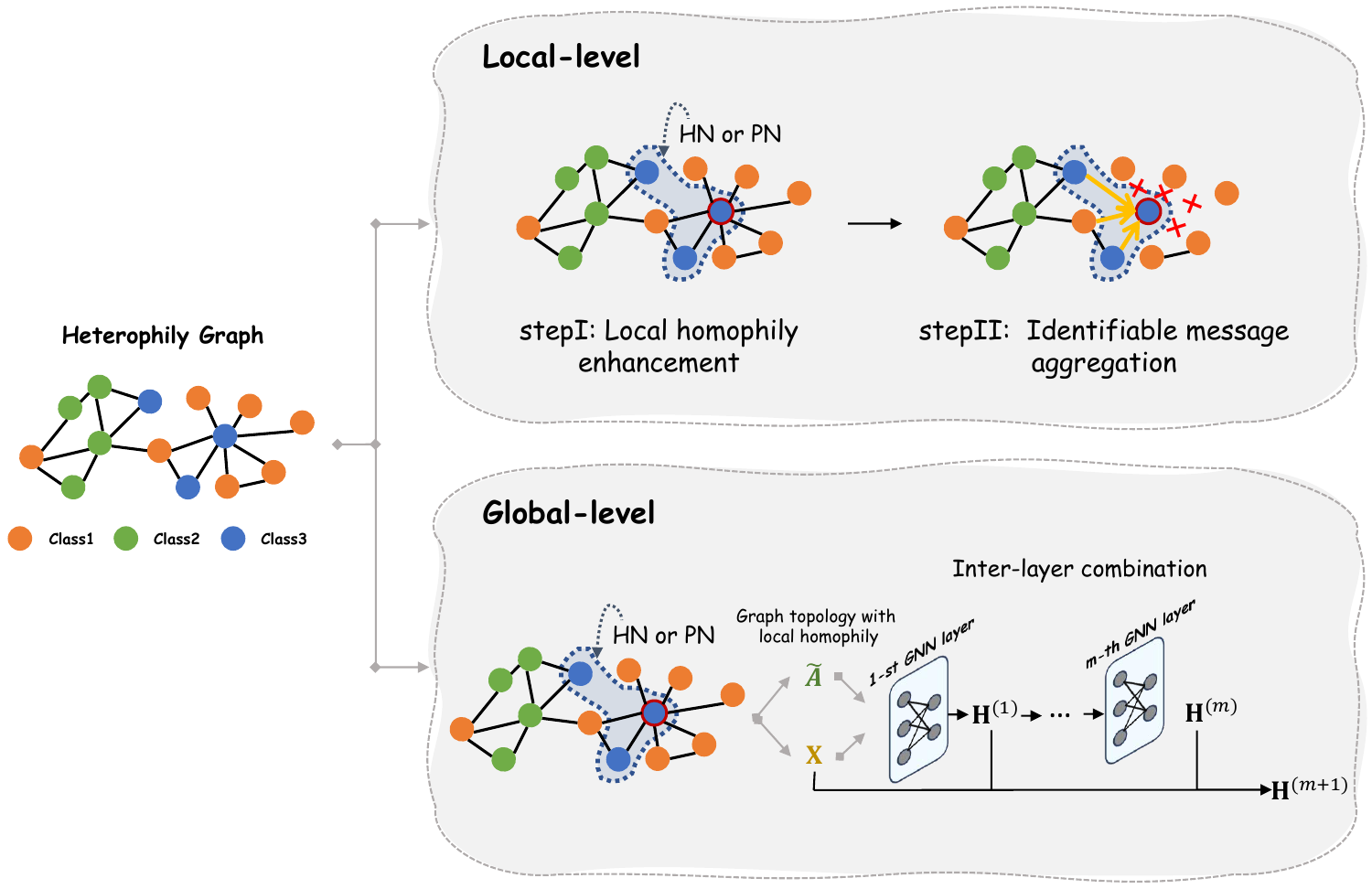}
\caption{A diagram of two types of hybrid methods: Local-level incorporation and global-level incorporation.}\label{fig:Hybrid2}
\end{figure}

\begin{table*}[t]
	\centering
    \caption{Summary of hybrid extension methods. \small{`HN' and `PN' mean `High-Order Neighbors Mixing' and `Potential Neighbors Discovery', respectively. `IA' and `IC' mean `Identifiable Message Aggregation' and `Inter-Layer Combination', respectively. `Local-Level' and `Global-Level' indicate whether the GNN architecture refinement method, incorporating neighbor extension technology, primarily focuses on heterophilic design within the message-passing layer or enhances effective inter-layer information transfer throughout the entire network architecture, respectively.}}
	\resizebox{0.85\textwidth}{!}{
\begin{tabular}{p{4cm}p{3.5cm}p{3.5cm}p{3.5cm}}
\toprule
\textbf{Publication Year and Venue} & \textbf{Method} & \textbf{Category} & \textbf{Hybrid Mode} \\ \midrule
{[}2020 NeurIPS{]}                               & H2GCN \cite{h2gcn_zhu2020beyond}           & HN \& IC \& IA    & Global-Level \\
{[}2021 KDD{]}                                   & WRGNN \cite{SureshBNLM21}           & PN \& IA          & Local-Level          \\
{[}2022 IJCAI{]}                                 & RAW-GNN \cite{jin2022raw}         & HN \& IA          & Local-Level          \\
{[}2022 TPAMI{]}                                 & NLGNN \cite{Liu2021NLGNN}           & PN \& IA          & Global-Level          \\
{[}2022 AAAI{]}                                  & BM-GCN \cite{bm_gcn_he2021block}          & PN \& IA          & Local-Level          \\
{[}2022 AAAI{]}                                  & Deformable GCN \cite{park2022deformable}  & PN \& IA          & Local-Level          \\
{[}2022 AAAI{]}                                  & GPNN \cite{yang2021graph}            & PN \& IC          & Local-Level                    \\ 
{[}2025 AAAI{]}& EG-GCN \cite{liu2025Intco} & PN \& IA    & Local-Level    \\\bottomrule
\end{tabular}
}
		\label{tab:main3}
\end{table*}
Recently, researchers have recognized that simultaneously expanding the non-local neighbor set and refining heterophily-guided GNN architectures, can significantly facilitate heterophilic graph representation learning. 
Generally, existing hybrid methods can be categorized into two groups: (1) local-level incorporation, which designs intra-layer heterophilic GNN message passing with the neighbor extension; and (2) global-level incorporation, which enhances effective inter-layer information transfer of heterophilic GNNs with the neighbor extension. The diagram of two types of hybrid methods is presented in Fig. \ref{fig:Hybrid2}.

The majority of methods fall into the category of local-level incorporation, typically following a two-step process.
First, they establish a graph topology that emphasizes local homophily, and then they design a message aggregation method based on this homophilic topology. For instance, WRGNN~\cite{SureshBNLM21} employs the degree sequence of neighbor nodes as a metric for measuring structural similarity between ego nodes. This is used to reconstruct a multi-relational graph that captures homophily in relational edges, followed by relational aggregation with explicit link weights. Additionally, BM-GCN~\cite{bm_gcn_he2021block} constructs a novel network topology using a block similarity matrix. This approach allows it to explore block-guided neighbors and perform classified aggregation with distinct aggregation rules for homophilic and heterophilic nodes. RAW-GNN~\cite{jin2022raw} employs breadth-first random walk searches to capture homophily information and depth-first searches to gather heterophily information. Instead of traditional neighborhoods, it utilizes path-based neighborhoods and introduces a new path-based aggregator based on Recurrent Neural Networks. GPNN~\cite{yang2021graph} leverages a pointer network to rank the potential neighbor nodes according to the attention scores or the relevant relationships to the ego node. In this way, potential neighbors that are most similar to the ego node in heterophilic graphs can be discovered and selected. Deformable GCN, as presented in the recent work by Park et al.~\cite{park2022deformable}, dynamically conducts convolution in multiple latent spaces, enabling it to capture both short and long-range dependencies between nodes. In this approach, the model also learns the positional embeddings (coordinates) of nodes to infer the relationships between nodes in an end-to-end manner. Depending on the position of a node, the convolution kernels are deformed using deformation vectors and make distinct transformations being applied to neighbors.

The global-level incorporation method, which is intended to establish effective homophilic guidance within the graph network framework. A typical example of such a method is H2GNN \cite{h2gcn_zhu2020beyond}, which incorporates a set of crucial design elements. These include the separation of ego- and neighbor-embedding, higher-order neighborhoods, and incorporation of intermediate representations into a graph neural network. These design choices collectively enhance the model's capacity to learn from the graph structure, especially when dealing with heterophilic relationships. Furthermore, NLGNN~\cite{Liu2021NLGNN} employs the attention mechanism to guide the sorting of non-local neighbors based on their importance. The entire architecture achieves efficient node classification on a heterogeneous graph using only two steps: straightforward attention-guided sorting and non-local aggregation.

\vspace{0mm}

\noindent\textbf{Significance.} Sections \ref{Non-Local} - \ref{Hybrid} discuss heterophilic GNNs, which are essential for applications like fraud detection \cite{ma2021comprehensive, gong2023beyond}, where anomalous nodes deliberately connect with benign ones to evade homophilic models (\eg, GCN). While heterophily-specific designs generally outperform homophilic GNNs, recent studies urge a re-evaluation of early empirical results. Specifically, \cite{platonov2022critical} identifies critical flaws (\eg, class imbalance, data leakage) in existing datasets and introduces rigorous new benchmarks. Furthermore, \cite{platonov2022characterizing} notes that traditional metrics hinder cross-dataset comparisons, proposing ``label informativeness" to transcend the simple homophily-heterophily dichotomy. Moreover, \cite{acm_luan2021heterophily,ma2022is} show that well-tuned GCNs can succeed on certain heterophilic graphs, delineating strict boundary conditions; \cite{ma2022is} explains that GCNs primarily fail when neighborhood distributions across distinct classes become indistinguishable. Collectively, these insights reflect a broader shift from empirical trial-and-error toward theoretically grounded design and robust generalization. Accordingly, we encourage evaluating future methods on reliable, high-quality datasets. Following this standard, Appendix C provides our additional comparisons.
\section{Discussion} \label{Discussion}
\subsection{Heterophily GNNs on Diverse Graph Types}

Current research predominantly focuses on heterophily modeling in single-relational static graphs, yet real-world applications (e.g., traffic flow dynamics and user-item hypergraph co-purchasing relationships \cite{luo2006exploring,ding2009pagerank,xiao2023hybrid}) typically involve diverse graph structures like dynamic graphs and hypergraphs. While recent studies \cite{zhou2022greto,lihypergraph,wang2022equivariant} have initiated concepts for heterophilic dynamic graphs and hypergraphs, this expansion also brings forth a set of unique challenges for heterophilic GNNs to address.

\subsubsection{Heterophily GNNs on Dynamic Graphs}
As a typical dynamic graph instantiation, spatial-temporal graph data is often inherently heterophilic. Zhou et al. \cite{zhou2022greto} recently aimed to validate this assertion by exploring spatial-temporal graphs in various real-world scenarios, such as traffic control, climate early-warning, and social networks. As depicted in Fig.~\ref{subfig:dyn}, observations in these scenarios tend to change over time, leading to dynamic and time-varying characteristics in the correlation between nodes. The challenge posed by these dynamic and time-varying characteristics is termed ``topology-task discordance'' in \cite{zhou2022greto}. This challenge arises when employing homophilic graph neural networks for node-level regression tasks on spatial-temporal graphs. It refers to the utilization of a pre-defined fixed topology for message passing on spatial-temporal graphs with diverse node-wise relations, resulting in the aggregation of neighbors that deviate from the intended target.

To formally measure spatial-temporal heterophily, two types of homophily measurements are introduced: ``intra-graph spatial homophily'', capturing node correlations within the same graph frames, and ``inter-graph transition homophily'' which extracts temporal evolution between adjacent temporal frames. The study further investigates the average homophily ratios for four real-world dynamic graphs: Metr-LA~\cite{li2018diffusion}, PeMS-Bay~\cite{li2018diffusion}, KnowAir~\cite{wang2020pm2}, and Temperature~\cite{wang2020pm2}. The homophily ratios within intra-graph frames and across temporally adjacent frames are observed to be low. This suggests that physically connected nodes may not necessarily share similar observations or exhibit the same directional variations.


Furthermore, adapting existing homophily theories to address the topology-task discordance based on the spatial-temporal characteristics of node-wise relationships remains a challenging endeavor. The primary hurdles can be summarized as follows:
\begin{itemize}
\item
Determining which pairs of nodes belong to homophily components in the absence of categorical labels.
\item
Incorporating target information to reconstruct node-wise correlations, thereby enabling improved and target-oriented aggregations.
\item
Leveraging dynamic local neighborhood environments for personalized high-order propagation.
\end{itemize}

These three aspects are expected to be key challenges in the study of heterophilic GNNs on dynamic graphs. This research direction will continue to demand the dedicated efforts of a significant number of researchers.

\subsubsection{Heterophily GNNs on Hypergraphs}
\begin{figure}[t!]
\centering
\subfigure[Spatial-temporal graph]{
    \includegraphics[width=0.18\textwidth]{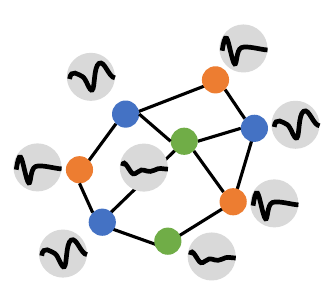}
    \label{subfig:dyn}
    }
\hspace{4mm}
\subfigure[Hypergraph]{
    \includegraphics[width=0.17\textwidth]{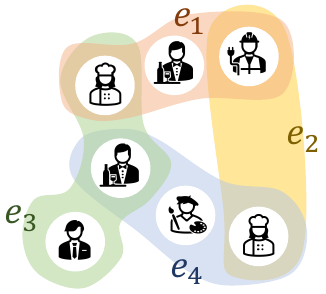}
    \label{subfig:hyp}
    }
\caption{Examples of (a) heterophilic dynamic graph and (b) heterophilic hypergraph.}\label{fig:DynamicGraphs}
\end{figure}

Fig. \ref{subfig:hyp} illustrates a heterophilic hypergraph using a social network as an example. Each node represents a social user, and the hyperedge connects all users in the same community. It is common for users within the same community to have different identity backgrounds but be connected by attributes such as interests and hobbies, resembling the concept of heterophily in graphs. Recently, heterophily has been demonstrated in \cite{veldt2021higher} to be a more prevalent phenomenon in hypergraphs compared to graphs. This is due to the challenge of expecting all nodes within a large hyperedge to share a common label. Li et al. \cite{lihypergraph} are the first to focus on heterophilic hypergraph learning (HHL), paving the way for HHL by addressing key gaps from multiple perspectives: measurement, dataset diversity, and baseline model development. Consequently, recent research on hypergraph neural network methods has shifted its focus towards highlighting their exceptional performance on heterophilic hypergraphs.

Inspired by hypergraph diffusion algorithms, ED-HNN \cite{wang2022equivariant} has been developed as a novel hypergraph neural network (HNN) architecture that can effectively approximate any continuous equivariant hypergraph diffusion operators. These operators have the capability to model a wide array of higher-order relations. In this research, it is asserted that predicting node labels in heterophilic hypergraphs is more intricate than in graphs since a hyperedge may consist of nodes from multiple categories. Therefore, the research further analyzes its learnable equivariant diffusion operator and demonstrates its superiority in predicting heterophilic node labels on four hypergraphs: Congress \cite{fowler2006legislative}, Senate \cite{fowler2006legislative}, Walmart \cite{amburg2020clustering}, and House \cite{chodrow2021hypergraph}. 

However, based on our review of related works, there is currently no established characterization or algorithmic design for specific heterophilic hypergraphs. The primary challenges revolve around accurately describing the heterophily of hypergraphs and developing architecture refinement methods that capture the local homophily of hypergraphs. This research direction is still promising but in its early stages of development.

\subsubsection{Heterophily GNNs on Heterogeneous Graphs}
Recent years have witnessed growing scholarly attention to heterophily characterization in heterogeneous graph analysis. Pioneering work by Lin et al. \cite{lin2024heterophily} made dual contributions: not only did they systematically demonstrate the coexistence of heterophilic patterns in homogeneous graphs and homophilic structures in heterogeneous graphs across real-world applications, but they also introduced the $\mathcal{H}^2GB$ benchmark dataset, effectively addressing critical gaps in standardized evaluation protocols. In parallel, Shen et al. \cite{shen2025heterophily} tackled heterophily challenges through an unsupervised perspective, formally defining the concept of semantic heterophily and developing LatGRL—a latent graph-guided reinforcement learning framework for hierarchical feature disentanglement. Despite these advancements, current research exhibits limitations, such as available benchmark datasets suffer from limited scale and scenario coverage. These unresolved issues necessitate coordinated efforts in developing theoretically grounded measurement systems, constructing large-scale multi-domain testbeds, and exploring cross-layer heterophily propagation mechanisms.

\begin{figure}[t!]
\centering
\includegraphics[width=0.49\textwidth]{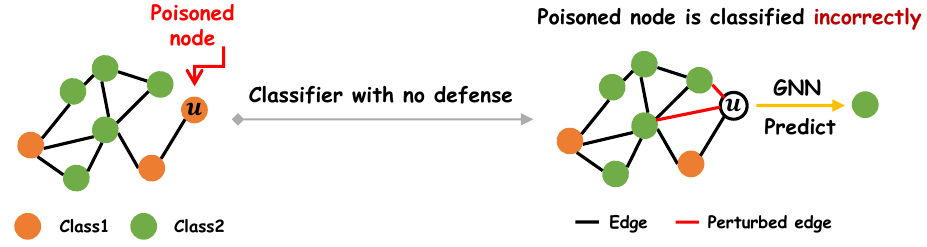}
\caption{Small, adversarial perturbations of the graph structure and node features lead GNN to misclassify target $u$.}\label{fig:ARGNN}
\end{figure}

\begin{figure*}[t!]
\centering
\includegraphics[width=\textwidth]{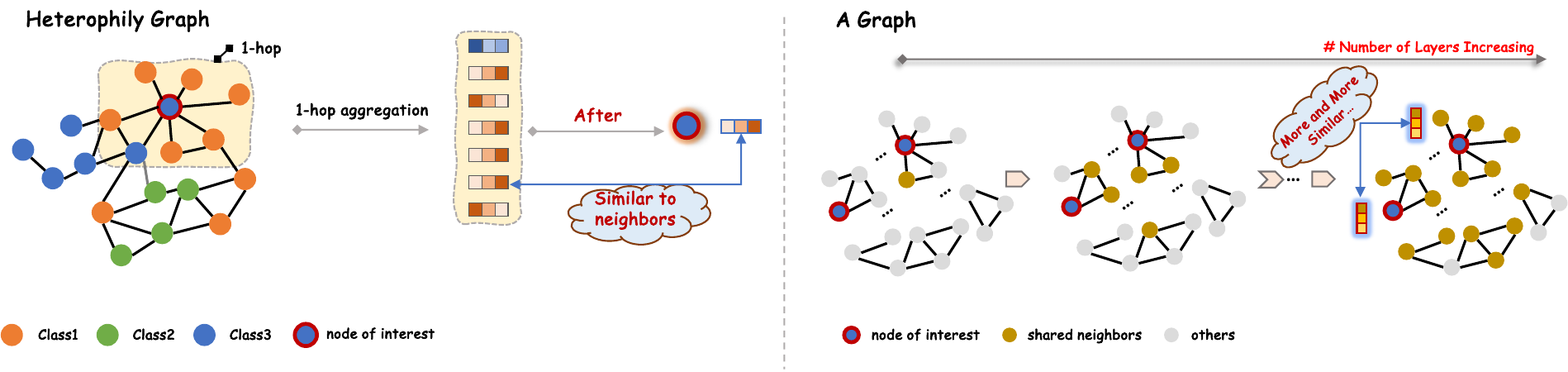}
\caption{Illustration of the connection between the heterophily property (Left) and the over-smoothing issue (Right) on GNNs.}\label{fig:hetero-vs-oversmoothing}
\end{figure*}

\subsection{Unveiling the Correlation between Graph Heterophily and Other Research Problems}
\subsubsection{Relation between Graph Heterophily and Adversarial Robustness}
Adversarial robustness in GNNs addresses model vulnerability to imperceptible structural perturbations, critical for trustworthy AI systems. While early studies focused on homophilic graphs \cite{sun2018adversarial,wu2019adversarial}, recent work recognizes structural attacks inherently exhibit heterophily (Fig. \ref{fig:ARGNN}), driving defense strategies:
\begin{itemize}
    \item GNNGUARD \cite{zhang2020gnnguard}: Detects feature-structure correlations to filter adversarial heterophilic edges, ensuring robust message propagation.
    \item GARNET \cite{deng2022garnet} \& ATDGIA \cite{chen2021understanding}: Treat heterophilic connections as adversarial edges, embedding anti-interference mechanisms.
    \item Theoretical Foundations \cite{zhu2022does}: Establishes homophily-heterophily dynamics under structural attacks, proving heterophily-aware designs inherently enhance robustness.
\end{itemize}
Current research converges on dual defense paradigms: explicit adversarial filtering and implicit robustness through heterophily-adaptive architectures.

Robustness and the interplay between homophily and heterophily in structural attacks on GNNs are key areas of investigation in these studies, with implications for improving robust GNN model designs.


\subsubsection{Relation between Graph Heterophily and Over-Smoothing}
Contemporary GNNs face dual challenges: heterophily violating homophily assumptions and over-smoothing eroding feature distinctiveness. As Fig.~\ref{fig:hetero-vs-oversmoothing} illustrates, these phenomena exhibit intrinsic connections - heterophilic aggregation leads to information dissipation in deep architectures \cite{pei2024multi}. Additionally, emerging studies reveal an intrinsic duality between these challenges: novel architectures designed for heterophilic graphs (e.g., FAGCN \cite{fagcn_bo2021beyond}, NLGNN \cite{Liu2021NLGNN}) exhibit inherent anti-over-smoothing properties, whereas certain over-smoothing mitigation strategies conversely enhance heterophilic learning \cite{chen2020simple}.

Theoretical unification reveals:
\begin{itemize}
    \item Sheaf-Theoretic Foundation: Cellular sheaf theory \cite{bodnar2022neural} demonstrated underlying sheaf structure of the graph is intimately linked with both heterophily and oversmoothing that influence the performance of GNNs.
    \item Signed Message Explanation: Yan et al. \cite{yan2021two} and Bodnar et al. \cite{bodnar2022neural} found evidence supporting the benefits of negatively signed edges in GNNs, albeit with different mathematical justifications. These theoretical findings also provide support for the effectiveness of technologies represented by FAGCN \cite{fagcn_bo2021beyond}.
\end{itemize}
These theoretical convergences enable unified frameworks achieving performance improvement across heterophilic benchmarks while maintaining stable higher layer performance - resolving the depth-performance trade-off in conventional GNNs.
\begin{figure}[t!]
\centering
\includegraphics[width=0.35\textwidth]{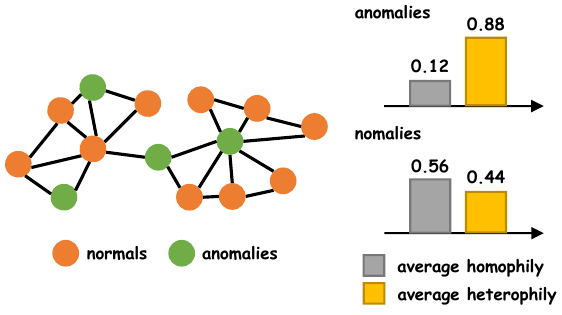}
\caption{A toy graph for anomaly detection. The average homophily and heterophily for anomalies and normals are presented.}\label{fig:FraudGNN}
\end{figure}

\subsubsection{Relation between Graph Heterophily and Graph Anomaly Detection}
Fraud detection, as a specialized anomaly detection task, increasingly focuses on heterophilic patterns in graph-structured financial data \cite{gao2023alleviating,shi2022h2,gong2023beyond,pan2025label}. As visualized in Fig. \ref{fig:FraudGNN}, fraudulent nodes typically form minority clusters surrounded by normal nodes, exhibiting higher heterophily ratios than legitimate entities \cite{gong2023beyond}.

Conventional graph anomaly detection (GAD) methods predominantly optimize homophilic message passing, inadvertently amplifying heterophily-induced noise during aggregation \cite{ma2021comprehensive}. This manifests as significant performance degradation in real-world fraud detection benchmarks. Recent advances address this through:
\begin{itemize}
    \item Structural Adaptation: Gong et al. \cite{gong2023beyond} develop graph sparsification techniques that selectively prune heterophilic edges while preserving homophilic dependencies.
    \item Distribution Alignment: Gao's SDS framework \cite{gao2023alleviating} mitigates structural distribution shifts between imbalanced classes via adversarial topology regularization.
    \item Spectral Decoupling: Hybrid approaches \cite{shi2022h2,gao2023addressing} dynamically combine low/high-frequency signals, achieving high precision in identifying heterophilic anomalies through adaptive graph filtering.
\end{itemize}
These methodologies establish heterophily suppression as critical for robust fraud detection systems, with spectral-graph fusion strategies demonstrating performance improvements over conventional GAD baselines.
\begin{figure}[t!]
\centering
\includegraphics[width=0.35\textwidth]{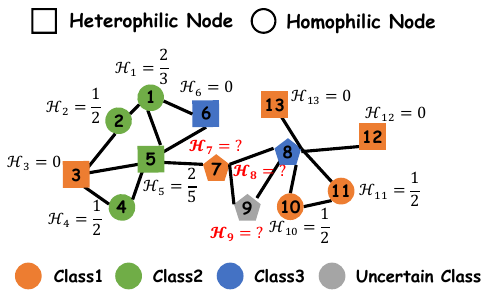}
\caption{Illustration of the uncertainty of connections in a graph.}\label{fig:Uncertainty_Graph}
\end{figure}
\subsubsection{Relation between Graph Heterophily and Uncertainty Modeling}
Uncertainty quantification in graph-structured data has become pivotal for reliable decision systems \cite{Tang2022Uncertainty,Liu2022Uncertainty,gao2023topology,Chen2023Uncertainty}. Under graph heterophily where nodes exhibit diverse connections (Fig. \ref{fig:Uncertainty_Graph}), topological uncertainty arises from heterogeneous relationship intensities that complicate information propagation.

Node-level analysis reveals real-world graphs often intermix homophilic/heterophilic connections, exposing GNNs' inherent bias toward homophilic patterns - models trained on semi-homophilic graphs show accuracy drops on purely heterophilic subsets \cite{Liu2022Uncertainty}. This bias-stability dilemma necessitates uncertainty-aware solutions:
\begin{itemize}
    \item Uncertainty Debiasing: Liu's framework \cite{Liu2022Uncertainty} employs epistemic uncertainty estimation to identify heterophilic nodes, preserving reliable predictions while rectifying high-uncertainty cases.
    \item Topology Augmentation: Gao et al. \cite{gao2023topology} propose GraphTU, a probabilistic method exploiting neighborhood variance to enhance minor class representation through non-parametric topology perturbation.
\end{itemize}
These advances establish heterophily-uncertainty interdependence as crucial for developing robust GNNs, offering new directions for social network analysis to biomedical discovery.

\noindent\textbf{Other Promising Heterophily Explorations. }Recently, researchers have sought to extend the design of heterophilic GNNs beyond the semi-supervised paradigm, focusing on self-supervised heterophilic GNNs \cite{he2023contrastive,li2023graphh,liu2023beyond}. 
It is noteworthy that He et al. \cite{he2023contrastive} identified commonalities between the establishment of the Graph Contrastive Learning (GCL) framework and the design of heterophilic GNNs. In the GCL framework, the task involves identifying additional positive samples belonging to the same class, while in the design of heterophilic GNNs, the objective is to identify other intra-class neighbors for each node. Recognizing this similarity, they integrated the GCL sampling strategy with homophily discrimination, creating a novel contrastive learning framework to address these crucial challenges. This framework draws inspiration from homophily, breaking traditional data enhancement strategies and providing a fresh perspective on establishing a GCL framework. Furthermore, it executes contrastive learning by sampling homophily neighbors as positive examples, offering innovative insights for the design of heterophilic GNNs. To address the limitations imposed by the strong homophily assumption and label dependency, Li et al. \cite{li2023graphh} introduced a novel multi-task model named PairE with a contrastive learning method. This model is designed to preserve information pertaining to both homophily and heterophily by transcending the localized node view. It achieves this by harnessing higher-level entities with more expressive power, enabling the representation of feature signals beyond homophily. In a similar vein and motivated by comparable considerations, Liu et al. \cite{liu2023beyond} propose an innovative self-supervised Graph Representation Learning method called Edge Heterophily Discriminating (GREET). By discriminating edges, GREET obtains contrastive views with both homophilic and heterophilic edges, leveraging them to learn expressive representations. 

\section{Future Directions} \label{Future}
GNNs for heterophilic graphs are fast developing in the past few years. Beyond current research, there still remain several open challenges and opportunities worthy of further attention and exploration. In this section, we discuss the following directions to stimulate future research.

\smallskip
\noindent\textbf{Interpretable Heterophilic GNNs.}
Interpretability is a crucial aspect of GNN models in risk-sensitive or privacy-related application fields, \eg, healthcare and cybersecurity. Although there are several studies on the interpretability \cite{ying2019gnnexplainer,pgexplainer_luo2020parameterized,agarwal2023evaluating} for homophilic GNNs, how to explain the predictions of heterophilic GNNs is still under-explored. Heterophily makes interpretability more challenging than homophily: since most local neighbor nodes are not in the same class as the ego nodes, extracting explainable subgraphs from highly heterophilic graph data is much harder, where both proximal and distant topological structures are required to be discovered and exploited. For instance, ShapeGGen~\cite{agarwal2023evaluating} typically proposed a dataset generator that can automatically generate a variety of benchmark datasets (\eg, varying graph sizes, degree distributions, homophilic and heterophilic graphs), accompanied by ground-truth explanations
From the data-centric view of heterophilic graphs, how to extract explainable subgraphs under complex similarity relationships between ego nodes and their potential neighbors on heterophilic graphs for interpretability is an open question. 
Moreover, from the model-centric view of heterophilic GNNs, how to explain the model's predictions on graphs with different degrees of heterophily also deserves further exploration.

\smallskip
\noindent\textbf{Scalable Heterophilic GNNs.} 
Current heterophilic GNNs are generally trained on relatively small graphs, which significantly limits their ability to model large-scale data and explore more complicated heterophilic patterns. Although possible solutions to tackle scalability can be borrowed from the mainstream graph sampling strategies ~\cite{chiang2019cluster,zeng2019graphsaint} for homophilic GNNs, the connections and relationships of heterophilic nodes would be undermined by sampling only mini-batches, especially when similar and dissimilar neighbors contribute differently to learning the ego-node representations. LINKX~\cite{lim2021large} recently verifies that even a simple MLP-based model could outperform GNNs for mini-batch training on large-scale heterophilic graphs. Moreover, HopGNN~\cite{chen2023node} introduces a hop interaction paradigm to address the scalability and over-smoothing problem of GNNs simultaneously, where its core idea is to convert the interaction target among nodes to pre-processed multi-hop features inside each node.
Hence, addressing the scalability problem of heterophilic graphs requires exploring more relationships between ego-neighbor node features and multi-hop information. How to keep the inherent heterophily unchanged when conducting sampling on heterophilic graphs is still an open question.

\smallskip
\noindent\textbf{Theoretical Heterophily Exploration.}
Despite the notable achievements of heterophilic GNNs across diverse tasks and datasets, their efficacy lacks a solid theoretical foundation to substantiate their impact on enhancing graph representation learning. Presently, many methods are predominantly designed based on intuition and evaluated through empirical experiments. Notably, a recent breakthrough by Ma et al. \cite{ma2022is} has unveiled that, under specific conditions, GCN can exhibit remarkable performance on heterophilic graphs. This discovery has been supported by comprehensive theoretical validation, outlining the distinctive characteristics and conditions under which heterophilic graphs can excel with GCN. The elucidation of these theoretical findings poses a significant challenge to the current rationale underlying the design of heterophilic GNNs. Consequently, there arises a critical need for intensified theoretical exploration in the future. This exploration should delve into aspects such as the impact of heterophily on altering the trends of generalization bounds. Bridging this explanatory gap is crucial for comprehending the extent to which heterophily influences the performance constraints of homophilic GNNs.

\smallskip
\noindent\textbf{Diverse Heterophily Learning Tasks.}
The research on heterophilic GNNs has primarily demonstrated success in node-level tasks, covering both semi-supervised \cite{abu2019mixhop,du2022gbk,kong2023goat} and unsupervised scenarios \cite{xiao2022decoupled,liu2023beyond,he2023contrastive}. A recent study by Zhou et al. \cite{zhou2022link} introduces a novel heterophilic framework named DisenLink, specifically designed for addressing link-level tasks on heterophilic graphs. In heterophilic graphs, DisenLink reveals that link formation is influenced by numerous latent factors, causing linked nodes to share similarity in certain factors while being dissimilar in others. This results in an overall low similarity between linked nodes. Many existing link prediction approaches operate under the homophily assumption, using similarity-based heuristics or representation learning methods to predict links. Therefore, accurately predicting links with heterophilic attributes, determined by specific potential similar factors, poses a significant challenge. DisenLink identifies the challenges of link-level tasks on heterophilic graphs, but there are still significant open questions in this domain. These include the development of reasonable heterophilic benchmark datasets and the refinement of GNN architectures for link-level tasks. Additionally, similar open questions exist for graph-level tasks, indicating that further exploration is needed in these areas.

\smallskip

\noindent\textbf{Broader Scope of Practical Applications.}
In the real world, heterophilic relationships are pervasive and exist widely in various graph-structured data \cite{du2024graph, wangmulti, wang2025cooperation, mele2022structural, choi2022finding, liu2024slgcn, maekawa2022beyond, chen2023drug}. However, current research on heterophilic GNNs predominantly focuses on specific strong heterophilic graphs, such as social networks \cite{mele2022structural} and web page networks \cite{choi2022finding}. Other fields, such as biomedicine \cite{liu2024slgcn} and chemistry \cite{maekawa2022beyond, chen2023drug}, which also naturally exhibit the heterophilic property, deserve more explorations.
For instance, in protein-protein interaction networks, proteins with interactions often belong to different gene ontologies. Moreover, in the field of drug discovery, DCMGCN~\cite{chen2022drug} verifies the heterophily property of the drug-drug networks and proposed a combination of intermediate representations, and high-similarity
neighborhoods, to boost GCN learning on the heterophily
and sparse drug-drug networks.
Besides, in the field of fraud detection, fraudsters tend to be more connected to regular users represented by fraud networks. And some existing research, \eg, DRAG~\cite{kim2023dynamic} and GAGA~\cite{wang2023label}, solved the problem by conducting the binary heterophilic graph classification task.
This highlights the untapped potential of heterophilic GNNs in a broader range of application areas. There is an expectation to extend the use of heterophilic GNNs to various fields, including financial networks, network security, community detection, biomedicine, and chemistry.


\section{Conclusion} \label{Conclusion}
This paper presents a comprehensive overview of graph neural networks for heterophilic graphs. We introduce a systematic taxonomy categorizing existing methods into three classes: non-local neighbor extension, GNN architecture refinement, and hybrid approaches. The study surveys current research progress and challenges in heterophilic graph learning, and discusses the relevance of graph heterophily to various domains such as model robustness, over-smoothing, and graph anomaly detection. In the end, we shared our insights into future research opportunities and directions that can contribute to the advancement of heterophilic GNNs.

\section*{Acknowledgments}
Shirui Pan was supported in part by the Australian Research Council (ARC) under grants FT210100097 and DP240101547. The work of Yixin Liu was partially supported by the Australian Research Council (ARC) Grant No. DE260101172. Yi Wang acknowledged the support from the Research Grants Council of Hong Kong (Project no. CityU 11301224). Ming Li acknowledged the support from National Natural Science Foundation of China (No. 62536006). Philip S. Yu is supported in part by NSF under grants III-2106758, and POSE-2346158.

\balance
\bibliographystyle{IEEEtran}
\bibliography{08reference}
\newpage
\newpage
\begin{appendices}

\section{Measure of Heterophily \& Homophily}

Table S-\ref{tab:Measures} summarizes the five homophily evaluation metrics. Note that the range of $\mathcal{H}_{node}$, $\mathcal{H}_{edge}$ and $\mathcal{H}_{class}$ is $[0,1]$. Graphs with strong homophily have higher $\mathcal{H}_{node}$, $\mathcal{H}_{edge}$ and $\mathcal{H}_{class}$ (closer to $1$); whereas graphs with strong heterophily have smaller $\mathcal{H}_{node}$, $\mathcal{H}_{edge}$ and $\mathcal{H}_{class}$ (closer to $0$). 
The measure $\mathcal{H}_{adj}$ and $\mathcal{H}_{un}$ are appropriately adjusted based on $\mathcal{H}_{edge}$. $\mathcal{H}_{adj}$ can occasionally have a negative value when $\mathcal{H}_{edge}$ is smaller than $\sum_{c=1}^C\left(\sum\limits_{v:y_v=c}|\mathcal{N}(v)|/(2|\mathcal{E}|)\right)^2$. Nevertheless, the overall trend of $\mathcal{H}_{adj}$ and $\mathcal{H}_{un}$ aligns with the previous three measures: larger values indicate stronger homophily, while smaller values suggest stronger heterophily.
Besides, we would like to emphasise that various metrics of homophily and heterophily are suited to specific use cases, and we aim to discuss several characteristics of these metrics.

For one thing, although the node homophily $\mathcal{H}_{node}$ and edge homophily $\mathcal{H}_{edge}$ are intuitive, they are relatively sensitive to the number of classes and the balance between these classes. 
This can make them difficult to interpret and render them incomparable across different datasets. 
For example, suppose that each node in a graph is connected to one node of each class. Then, both edge homophily and node homophily for this graph will be equal to $1/C$. As a result, these metrics will produce widely different values for graphs with different numbers of classes, despite these graphs being similar in exhibiting no homophily. For another thing, class homophily does not account for variations in node degrees when correcting the fraction of intra-class edges by $n_c/|\mathcal{V}|$. Specifically, if nodes of class $c$ have, on average, larger degrees than $2|\mathcal{E}|/|\mathcal{V}|$, then the probability that a random edge goes to that class can be significantly larger than $n_c/|\mathcal{V}|$. Secondly, class homophily only considers positive deviations from $n_c/|\mathcal{V}|$, meaning that it doesn't take into account classes with heterophilic connectivity patterns. As advised by \cite{platonov2022characterizing} and \cite{mironovrevisiting}, $\mathcal{H}_{adj}$ and $\mathcal{H}_{un}$ are better alternative to the commonly used measures, as it satisfies most of the desirable properties.

\begin{table*}[h]
	\centering
    \caption{Summary of existing measure methods of heterophily and homophily.}
	\resizebox{0.98\textwidth}{!}{
\begin{tabular}{m{3.5cm} m{6.8cm} m{8cm}}
\toprule[1.5pt]
\textbf{Measure Methods}   & \textbf{Definition}    & \textbf{Description} \\ \midrule[0.8pt]
Node Homophily             & $\mathcal{H}_{node} = \frac{1}{|\mathcal{V}|} \sum_{v \in \mathcal{V}} \frac{|\{u \in \mathcal{N}(v): y_v = y_u \}|}{|\mathcal{N}(v)|}$                & \multirow{2}{*}{
\makecell[c{>{\parindent 0em}p{8cm}}]{Both $\mathcal{H}_{node}$ and $\mathcal{H}_{edge}$ are within the range of $[0,1]$. These two measures are intuitive, but they are relatively sensitive to the number of classes and the balance between these classes.}} \\ \cmidrule{0-1}
Edge Homophily             & $\mathcal{H}_{edge} = \frac{|\{(v, u) \in \mathcal{E}: y_{v}=y_{u}\}|}{|\mathcal{E}|}$     &  \\\midrule
Class Homophily            & $\mathcal{H}_{class} = \frac{1}{C-1}\sum\limits_{c=1}^C\left[\frac{\sum\limits_{v:y_v=c}|\{u \in \mathcal{N}(v): y_v = y_u \}|}{\sum_{v:y_v=c}|\mathcal{N}(v)|}-\frac{n_c}{|\mathcal{V}|}\right]_{+}$                    & $\mathcal{H}_{class}$ avoids the class sensitivity issue of $\mathcal{H}_{node}$ and $\mathcal{H}_{edge}$ and falls within the range $[0,1]$. However, it doesn't consider certain desirable properties of a reasonable homophily measure, such as variations in node degrees. \\ \midrule
Adjusted Homophily         & $\mathcal{H}_{adj} = \frac{\mathcal{H}_{edge}-\sum_{c=1}^C\left(\sum\limits_{v:y_v=c}|\mathcal{N}(v)|/(2|\mathcal{E}|)\right)^2}{1-\sum_{c=1}^C\left(\sum\limits_{v:y_v=c}|\mathcal{N}(v)|/(2|\mathcal{E}|)\right)^2}$                    & $\mathcal{H}_{adj}$ is comparable across different datasets with varying numbers of classes and class size balances. In general, $\mathcal{H}_{adj}< \mathcal{H}_{edge}$ holds, and $\mathcal{H}_{adj}$ sometimes take on negative values.\\ \midrule
Unbiased Homophily & $\mathcal{H}_{un}=\frac{\sum_{i<j}(\sqrt{c_{ii}c_{jj}}-c_{ij})}{\sum_{i<j}(\sqrt{c_{ii}c_{jj}}+c_{ij})}$                    & $\mathcal{H}_{un}$ is designed for reliable application across differing label distributions that satisfies the vast majority of these properties.\\ \bottomrule[1.5pt]
\end{tabular}}
\label{tab:Measures}
\end{table*}

\section{Real-World Benchmarks}
To provide the data support for the development of heterophilic graph learning, we collect and list current real-world benchmarks of heterophilic graph datasets with detailed statistics, as shown in Tables S-~\ref{tab:datasets1} - \ref{tab:datasets3}. All datasets can be divided in three categories: standard datasets, large-scale datasets and high-quality datasets.
\begin{table*}[h]
\centering

\caption{Statistics of standard heterophilic graph benchmarks. }
\resizebox{0.95\textwidth}{!}{
\begin{tabular}{p{4.5cm}p{3cm}cccccccccc}
\toprule
Types       & Datasets      & \# Nodes  & \# Edges    & \# Features & \# Classes & $\mathcal{H}_{node}$ & $\mathcal{H}_{edge}$  & $\mathcal{H}_{class}$ & $\mathcal{H}_{adj}$ & $\mathcal{H}_{un}$ \\ \midrule
\multirow{3}{*}{WebKB Webpage} & Cornell & 183 & 295 & 1,703 & 5 & 0.11 & 0.30 & 0.06 & -0.01 & -0.08\\
& Texas & 183 & 325 & 1,703 & 5 & 0.06 & 0.11 & 0.15 & -0.22 & -0.53\\
& Wisconsin & 251 & 499 & 1,703 & 5 & 0.16 & 0.17 & 0.16 & -0.10 & -0.31\\\midrule
Author Co-occurrence & Actor & 7,600 & 29,926 & 932 & 5 & 0.24 & 0.22 & 0.01 & 0.06 & 0.01 \\\midrule
\multirow{2}{*}{Wikipedia Webpage} & Chameleon & 2,277 & 36,051 & 2,325 & 5 & 0.25 & 0.23 & 0.05 & 0.09 & 0.09 \\
& Squirrel & 5,201 & 216,933 & 2,089 & 5 & 0.22 & 0.22 & 0.05 & 0.05 & 0.04\\ \bottomrule
\end{tabular}}
\label{tab:datasets1}
\end{table*}

\subsection{Standard Datasets}
Most current works conduct experiments for empirical evaluations on the first $6$ datasets in Table S-~\ref{tab:datasets1}, which are originally presented by the work \cite{geom_pei2020geom}. Specifically, Cornell, Texas, and Wisconsin are three subdatasets from WebKB web-page dataset \cite{garcia2016using}, where nodes are the web pages from computer science departments of different universities, and edges are mutual links between pages. Chameleon and Squirrel are two web page datasets collected from Wikipedia \cite{rozemberczki2021multi}, where nodes are web pages on specific topics and edges are hyperlinks between them. Actor is an actor co-occurrence network \cite{tang2009social}, where nodes are actors and edges mean two actors are co-occurred on the same Wikipedia page. 

\subsection{Large-Scale Datasets}
Subsequently, a series of new benchmark datasets with larger scales from diverse areas are collected and released by \cite{lim2021large} and~\cite{lim2021new}, including 
two citation network datasets (ArXiv-Year and Snap-Patents), five online social network datasets (Penn94, Pokec, Genius, Deezer-Europe, and Twitch-Gamers), as well as a web-page hotel and restaurant review dataset (YelpChi). The statistics of large-scale heterophilic graph benchmarks are summuarized in Table S-\ref{tab:datasets2}.
\begin{table*}[h]
\centering
\caption{Statistics of large-scale heterophilic graph benchmarks. }
\resizebox{0.95\textwidth}{!}{
\begin{tabular}{p{3.5cm}p{2.5cm}cccccccccc}
\toprule
Types       & Datasets      & \# Nodes  & \# Edges    & \# Features & \# Classes & $\mathcal{H}_{node}$ & $\mathcal{H}_{edge}$ & $\mathcal{H}_{class}$ & $\mathcal{H}_{adj}$ & $\mathcal{H}_{un}$\\ \midrule
\multirow{2}{*}{Citation}          & ArXiv-Year    & 169,343   & 1,166,243   & 128           & 5        & 0.29               & 0.22      & 0.07    & 0.01   & -0.05\\
                                   & Snap-Patents  & 2,923,922 & 13,975,791  & 269           & 5        & 0.19               & 0.22         & 0.08    & 0.00     & -0.08\\\midrule
\multirow{5}{*}{Social Networks}   & Deezer-Europe & 28,281    & 92,752      & 31,241      & 2          & 0.53          & 0.53   & 0.03    & 0.03    & 0.03\\
                                   & Penn94        & 41,554    & 1,362,229   & 5           & 2          & 0.48               & 0.47         & 0.07    & 0.40    & 0.03\\
                                   & Twitch-Gamers & 168,114   & 6,797,557   & 7           & 2          & 0.56               & 0.55         & 0.09    & 0.09    & 0.09\\
                                   & Genius        & 421,961   & 984,979     & 12           & 2         & 0.10               & 0.62         & 0.10    & 0.17     &  -0.07 \\
                                   & Pokec         & 1,632,803 & 30,622,564  & 65           & 2         & 0.39               & 0.45         & 0.08    & 0.24    & -0.11 \\\midrule
Webpage Review                     & YelpChi       & 45,954    & 3,846,979   & 32          & 2          & 0.77           & 0.77      & 0.05    & 0.72    & 0.10\\ \bottomrule
\end{tabular}
}
\label{tab:datasets2}
\end{table*}

\begin{table*}[!b]
\centering
\caption{Statistics of high-quality heterophilic graph benchmarks. }
\resizebox{0.95\textwidth}{!}{
\begin{tabular}{p{3cm}p{2.5cm}cccccccccc}
\toprule
Types       & Datasets      & \# Nodes  & \# Edges    & \# Features & \# Classes & $\mathcal{H}_{node}$ & $\mathcal{H}_{edge}$ & $\mathcal{H}_{class}$ & $\mathcal{H}_{adj}$ & $\mathcal{H}_{un}$\\ \midrule
\multirow{1}{*}{Wikipedia Webpage} & Roman-Empire        & 22,662 & 32,927 & 300           & 18     & 0.05    & 0.05     & 0.02         & -0.05            & -0.49\\ \midrule
\multirow{1}{*}{Product Co-Purchasing} & Amazon-Ratings      & 24,492 &  93,050 & 300           & 5   & 0.38      & 0.38           & 0.13       & 0.14         & 0.28 \\ \midrule
\multirow{1}{*}{Game} & Minesweeper          & 10,000 &  39,402 & 7           & 2   & 0.68      & 0.68      & 0.01            & 0.01    &  0.01  \\ \midrule
\multirow{2}{*}{Social Networks} & Tolokers          & 11,758 & 519,000 & 10           & 2  & 0.63       & 0.59      & 0.18            & 0.09           &  0.10 \\
& Questions          & 48,921 & 153,540 & 301           & 2  & 0.90       & 0.84     &  0.08            & 0.02       &  0.06  \\\bottomrule
\end{tabular}}
\label{tab:datasets3}
\end{table*}

\begin{table*}[!b]
\centering
\caption{Performance Comparison on Five High-Quality Heterophilic Graphs. }
\resizebox{0.95\textwidth}{!}{
\begin{tabular}{c|ccc|cccccc}
\toprule
\multirow{2}{*}{Datasets} & \multicolumn{3}{c|}{Traditional GNNs}           & \multicolumn{6}{c}{Heterophilic GNNs}    \\
& GCN              & SAGE         & GAT          & H2GCN        & GloGNN  & LINKX            & ACM-GCN          & AGS-GNN           & PD-GNN       \\ \midrule
roman-empire              & 73.69 $\pm$ 0.74 & 85.74 $\pm$ 0.67 & 80.87 $\pm$ 0.30 & 60.11 $\pm$ 0.52 & 59.63 $\pm$ 0.69 & 59.14 $\pm$ 0.45 & 71.42 $\pm$ 0.38 & 80.49 $\pm$ 0.48 & 89.23 $\pm$ 0.56 \\
amazon-ratings            & 48.70 $\pm$ 0.63 & 53.63 $\pm$ 0.39 & 49.09 $\pm$ 0.63 & 36.47 $\pm$ 0.23 & 36.89 $\pm$ 0.14 & 52.68 $\pm$ 0.26 & 52.94 $\pm$ 0.23 & 53.21 $\pm$ 0.46 & 50.96 $\pm$ 0.43 \\
minesweeper               & 89.75 $\pm$ 0.52     & 93.51 $\pm$ 0.57 & 92.01 $\pm$ 0.68 & 89.71 $\pm$ 0.31 & 51.08 $\pm$ 1.23 & 80.02 $\pm$ 0.03 &  90.47 $\pm$ 0.57 & 85.56 $\pm$ 0.28 & 94.03 $\pm$ 0.45 \\
tolokers                  & 83.64 $\pm$ 0.67     & 82.43 $\pm$ 0.44 & 83.70 $\pm$ 0.47 & 73.35 $\pm$ 1.01 & 73.39 $\pm$ 1.17 & 80.07 $\pm$ 0.53 & 80.45 $\pm$ 0.54 & 80.52 $\pm$ 0.41 & 84.83 $\pm$ 0.40 \\
questions                 & 76.09 $\pm$ 1.27     & 76.44 $\pm$ 0.62 & 77.43 $\pm$ 1.20 & 63.59 $\pm$ 1.46 & 65.74 $\pm$ 1.19 & 97.06 $\pm$ 0.03 & 97.02 $\pm$ 1.67 & 97.27 $\pm$ 0.04 & 78.88 $\pm$ 0.94  \\\bottomrule
\end{tabular}}
\label{tab:benchmarks}
\end{table*}

\begin{table*}[h]
	\centering
    \caption{A summary of open-source implementations.}
	\resizebox{\textwidth}{!}{
\begin{tabular}{lll}
\toprule
\textbf{Method}                                           & \textbf{Framework} & \textbf{Github Link}                                       \\ \midrule
\multirow{2}{*}{JK-Net (2018) \cite{xu2018representation}}             & TensorFlow                  & \href{https://github.com/ShinKyuY/Representation_Learning_on_Graphs_with_Jumping_Knowledge_Networks}{https://github.com/ShinKyuY/Representation\underline{~}Learning\underline{~}on\underline{~}Graphs\underline{~}with\underline{~}Jumping\underline{~}Knowledge\underline{~}Networks}              \\
& PyTorch      &  \href{https://github.com/xnuohz/JKNet-dgl}{https://github.com/xnuohz/JKNet-dgl}          \\
MixHop (2019) \cite{abu2019mixhop}          & PyTorch             &     \href{https://github.com/samihaija/mixhop}{https://github.com/samihaija/mixhop}        \\ 
GCNII (2020) \cite{chen2020simple}          & PyTorch             &     \href{https://github.com/chennnM/GCNII}{https://github.com/chennnM/GCNII}        \\ 
Geom-GCN (2020) \cite{geom_pei2020geom}          & PyTorch             &     \href{https://github.com/alexfanjn/GeomGCN_PyG}{https://github.com/alexfanjn/GeomGCN\underline{~}PyG}        \\ 
GPR-GNN (2020) \cite{chien2020adaptive}          & PyTorch             &     \href{https://github.com/jianhao2016/GPRGNN}{https://github.com/jianhao2016/GPRGNN}        \\  
H2GCN (2020) \cite{h2gcn_zhu2020beyond}          & PyTorch             &     \href{https://github.com/GemsLab/H2GCN}{https://github.com/GemsLab/H2GCN}        \\ 
TDGNN (2021) \cite{wang2021tree}          & PyTorch             &     \href{https://github.com/Leo-Q-316/TDGNN}{https://github.com/Leo-Q-316/TDGNN}        \\ 
SimP-GCN (2021) \cite{simp_jin2021node}          & PyTorch             &     \href{https://github.com/ChandlerBang/SimP-GCN}{https://github.com/ChandlerBang/SimP-GCN}        \\ 
U-GCN (2021) \cite{jin2021universal}          & PyTorch             &     \href{https://github.com/jindi-tju/U-GCN}{https://github.com/jindi-tju/U-GCN}        \\ 
FAGCN (2021) \cite{fagcn_bo2021beyond}          & PyTorch             &     \href{https://github.com/bdy9527/FAGCN}{https://github.com/bdy9527/FAGCN}        \\ 
CPGNN (2021) \cite{zhu2021graph}          & TensorFlow             &     \href{https://github.com/GemsLab/CPGNN}{https://github.com/GemsLab/CPGNN}        \\
CGCN (2021) \cite{yan2021two}          & PyTorch             &     \href{https://github.com/Yujun-Yan/Heterophily_and_oversmoothing}{https://github.com/Yujun-Yan/Heterophily\underline{~}and\underline{~}oversmoothing}        \\ 
ACM-GNN (2022) \cite{acm_luan2021heterophily}          & PyTorch             &     \href{https://github.com/SitaoLuan/ACM-GNN}{https://github.com/SitaoLuan/ACM-GNN}        \\
EvenNet (2022) \cite{lei2022evennet}    & PyTorch              & \href{https://github.com/Leirunlin/EvenNet}{https://github.com/Leirunlin/EvenNet}                       \\
Diag-NSD (2022) \cite{bodnar2022neural} & PyTorch              & \href{https://github.com/twitter-research/neural-sheaf-diffusion}{https://github.com/twitter-research/neural-sheaf-diffusion} \\
MPNN (2022) \cite{huang2022local}       & PyTorch              & \href{https://github.com/nhuang37/spectral-inspired-gnn}{https://github.com/nhuang37/spectral-inspired-gnn}          \\
GARNET (2022) \cite{deng2022garnet}     & PyTorch              & \href{https://github.com/cornell-zhang/GARNET}{https://github.com/cornell-zhang/GARNET}                    \\
LW-GNN (2022) \cite{dai2022labelwise}   & PyTorch              &       \href{https://github.com/EnyanDai/LWGCN}{https://github.com/EnyanDai/LWGCN}                          \\
GBK-GNN (2022) \cite{du2022gbk}         & PyTorch              &                       \href{https://github.com/Xzh0u/GBK-GNN}{https://github.com/Xzh0u/GBK-GNN}                           \\
F2GNN (2022) \cite{wei2022designing}    & PyTorch              & \href{https://github.com/LARS-research/F2GNN}{https://github.com/LARS-research/F2GNN}                      \\
GloGNN (2022) \cite{li2022finding}      & PyTorch              &           \href{https://github.com/RecklessRonan/GloGNN}{https://github.com/RecklessRonan/GloGNN}   \\
GIND (2022) \cite{chen2022optimization} & PyTorch              &           \href{https://github.com/7qchen/GIND}{https://github.com/7qchen/GIND}  \\
BM-GCN (2022) \cite{bm_gcn_he2021block}        & PyTorch              &    \href{https://github.com/hedongxiao-tju/BM-GCN}{https://github.com/hedongxiao-tju/BM-GCN}    \\
Deformable GCN (2022) \cite{park2022deformable}& PyTorch              &    \href{https://github.com/mlvlab/DeformableGCN}{https://github.com/mlvlab/DeformableGCN}             \\
GIA-HAO (2022) \cite{chen2022hao}              & PyTorch             &     \href{https://github.com/LFhase/GIA-HAO}{https://github.com/LFhase/GIA-HAO}          \\
HeteRobust (2022) \cite{zhu2021graph}          & PyTorch             &     \href{https://github.com/GemsLab/HeteRobust}{https://github.com/GemsLab/HeteRobust}        \\      
GREET (2023) \cite{liu2023beyond}              & PyTorch             &     \href{https://github.com/yixinliu233/GREET}{https://github.com/yixinliu233/GREET}        \\   
HopGNN (2023) \cite{chen2023node}              & PyTorch             &     \href{https://github.com/JC-202/HopGNN}{https://github.com/JC-202/HopGNN}        \\ 
ED-HNN (2023) \cite{wang2022equivariant}       & PyTorch             &     \href{https://github.com/Graph-COM/ED-HNN}{https://github.com/Graph-COM/ED-HNN}        \\ 
Ordered GNN (2023) \cite{song2022ordered}      & PyTorch             &     \href{https://github.com/LUMIA-Group/OrderedGNN}{https://github.com/LUMIA-Group/OrderedGNN}        \\ 
GReTo (2023) \cite{zhou2022greto}          & PyTorch             &     \href{https://github.com/zzyy0929/ICLR23-GReTo}{https://github.com/zzyy0929/ICLR23-GReTo}        \\ 
SNGNN (2023) \cite{zou2023similarity}          & PyTorch             &     \href{https://github.com/MinhZou/SNGNN}{https://github.com/MinhZou/SNGNN}        \\ 
CAGNNs (2023) \cite{chenexploiting}          & PyTorch             &     \href{https://github.com/JC-202/CAGNN}{https://github.com/JC-202/CAGNN}        \\ 
CGP (2023) \cite{liu2023comprehensive}          & PyTorch             &     \href{https://github.com/LiuChuang0059/CGP}{https://github.com/LiuChuang0059/CGP}        \\ 
RFA-GNN (2023) \cite{wu2023beyond}          & PyTorch             &     \href{https://github.com/LirongWu/RFA-GNN}{https://github.com/LirongWu/RFA-GNN}        \\ 
HES-GSL (2023) \cite{wu2023homophily}          & PyTorch             &     \href{https://github.com/LirongWu/Homophily-Enhanced-Self-supervision}{https://github.com/LirongWu/Homophily-Enhanced-Self-supervision}        \\ 
GDN (2023) \cite{gao2023alleviating}          & PyTorch             &     \href{https://github.com/blacksingular/wsdm_GDN}{https://github.com/blacksingular/wsdm\underline{~}GDN}        \\ 
SE-GSL (2023) \cite{zou2023se}          & PyTorch             &     \href{https://github.com/RingBDStack/SE-GSL}{https://github.com/RingBDStack/SE-GSL}        \\ 
GHRN (2023) \cite{gao2023addressing}          & PyTorch             &     \href{https://github.com/blacksingular/GHRN}{https://github.com/blacksingular/GHRN}        \\ 
AutoGCN (2023) \cite{wu2023beyond2}          & PyTorch             &     \href{https://github.com/nnzhan/AutoGCN}{https://github.com/nnzhan/AutoGCN}        \\ 
GOAL (2023) \cite{zheng2023finding}          & PyTorch             &     \href{https://github.com/zyzisastudyreallyhardguy/GOAL-Graph-Complementary-Learning}{https://github.com/zyzisastudyreallyhardguy/GOAL-Graph-Complementary-Learning}        \\ 
L2A (2023) \cite{wu2023learning}          & PyTorch             &     \href{https://github.com/LirongWu/L2A}{https://github.com/LirongWu/L2A}        \\
GPRGNN-L (2023) \cite{wu2023leveraging}          & PyTorch             &     \href{https://github.com/lucio-win/PKDD2023}{https://github.com/lucio-win/PKDD2023}        \\ 
PC-Conv (2024) \cite{li2024pc}          & PyTorch             &     \href{https://github.com/uestclbh/PC-Conv}{https://github.com/uestclbh/PC-Conv}        \\ 
TFE-GNN (2024) \cite{duan2024unifying}          & PyTorch             &     \href{https://github.com/graphNN/TFEGNN}{https://github.com/graphNN/TFEGNN}        \\ 
PEGFAN (2024) \cite{li2024permutation}          & PyTorch             &     \href{https://github.com/zrgcityu/PEGFAN}{https://github.com/zrgcityu/PEGFAN}        \\
AAGCN (2025) \cite{rey2025redesigning}          & PyTorch             &     \href{https://github.com/reysam93/adaptive_agg_gcn}{https://github.com/reysam93/adaptive\_agg\_gcn}    \\      GSF-GNN (2025)     \cite{liu2025global}          & PyTorch             &     \href{https://github.com/huijieliu2023/GSF-GNN}{https://github.com/huijieliu2023/GSF-GNN}  \\
PathLens (2025)     \cite{goyal2025pathlens}          & PyTorch             &     \href{https://github.com/goyalkaraniit/PathLens}{https://github.com/goyalkaraniit/PathLens}  \\
$\mathcal{H}^{2}$GB (2025)     \cite{lin2025heterophily}          & PyTorch             &     \href{https://github.com/junhongmit/H2GB/}{https://github.com/junhongmit/
H2GB/}  \\
DiRW (2025)     \cite{su2025dirw}          & PyTorch             &     \href{https://github.com/dhsiuu/DiRW}{https://github.com/dhsiuu/DiRW}  \\
RASH (2025)     \cite{zheng2025enhancing}          & PyTorch             &     \href{https://github.com/zhengziyu77/RASH}{https://github.com/zhengziyu77/RASH} 
\\
\bottomrule
\end{tabular}
}
		\label{tab:Implementation}
\end{table*}
\end{appendices}

\subsection{High-Quality Datasets}
Recent research \cite{platonov2022critical} has highlighted significant issues with standard datasets used to evaluate heterophily-specific models. These issues make results obtained using such datasets unreliable. One of the most notable drawbacks is the presence of a large number of duplicate nodes in datasets like squirrel and chameleon, which results in train-test data leakage. Removing these duplicate nodes has a strong impact on GNN performance in these datasets. To address these challenges, a set of reliable heterophilic graphs with diverse properties has been proposed. These graphs include Roman-Empire, Amazon-Ratings, Minesweeper, Tolokers, and Questions. This effort aims to create a collection of trustworthy heterophilic benchmarks that can support and advance research and applications of GNNs on heterophilic graphs. It is expected that more publicly available benchmarks for heterophily will be developed to further facilitate research in this area. The statistics of high-quality heterophilic graph benchmarks are summuarized in Table S-\ref{tab:datasets3}.

\section{Performance Comparison}
Table \ref{tab:benchmarks} presents a performance comparison between traditional GNNs and heterophily-specific models across five high-quality benchmark datasets. Many heterophily-specific GNNs originally reported results on conventional standard datasets, their evaluation on these more rigorous benchmarks provides a more realistic assessment of their utility. The results indicate that although some early specialized models are occasionally surpassed by well-tuned traditional GNNs, recent state-of-the-art architectures—such as ACM-GCN \cite{acm_luan2021heterophily}, AGS-GNN \cite{das2024ags}, and PD-GNN \cite{lu2025flexible}—consistently outperform traditional baselines. This empirical evidence validates that heterophily-specific inductive biases are indeed effective, demonstrating significant practical efficacy across diverse and challenging graph structures.

\section{Open-Source Implementations}
We have gathered the implementations of heterophilic GNN approaches discussed in this survey if open-source code for these methods is available. You can find the hyperlinks to the source codes in Table S-\ref{tab:Implementation}.

\begin{figure*}[!t]
\centering
\includegraphics[width=0.55\textwidth]{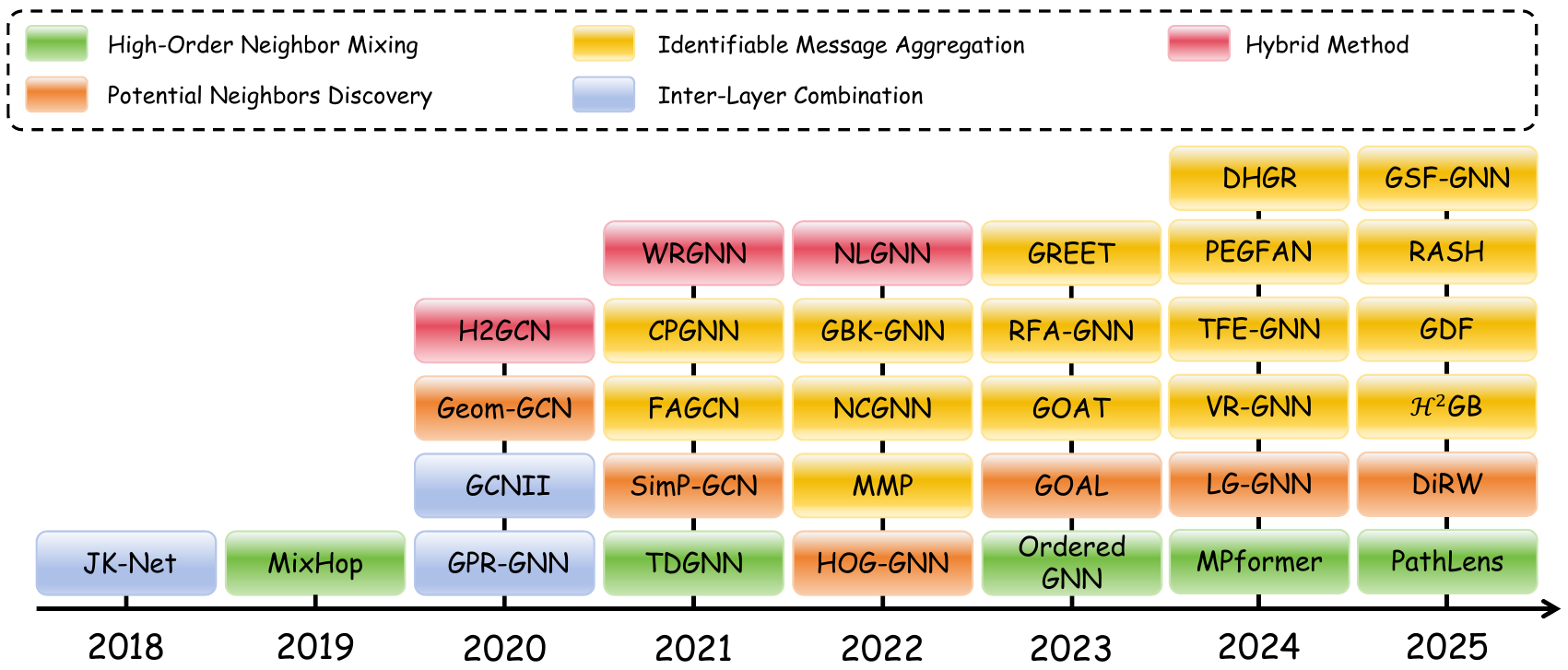}
\includegraphics[width=0.4\textwidth]{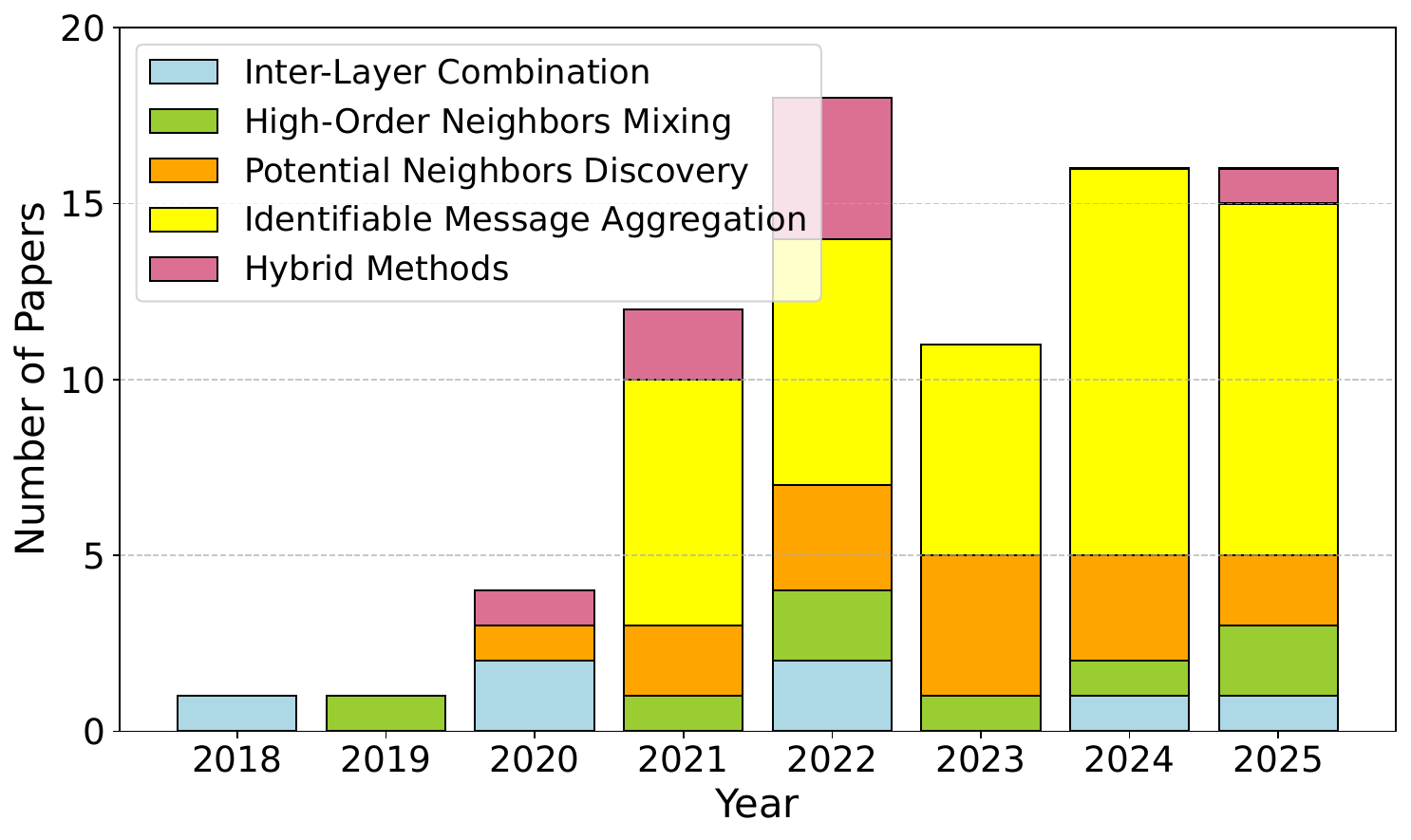}
\caption{A timeline of heterophilic GNNs development.}\label{fig:Timeline}
\end{figure*}


\section{Timeline of Heterophily GNNs}
Figure S\ref{fig:Timeline} presents two corresponding timelines: the left illustrates some of the most notable heterophilic GNNs developed since 2018, while the right displays the evolution view of these technical works summarized in this survey. Early works in the field include straightforward layer-inter combination methods like JK-Net \cite{xu2018representation}, GPR-GNN \cite{chien2020adaptive}, and GCNII \cite{chen2020simple}. Another category of early works explores non-local neighbor mixing and discovery intuitively, exemplified by MixHop \cite{abu2019mixhop} and Geom-GCN \cite{geom_pei2020geom}. H2GCN \cite{h2gcn_zhu2020beyond} represents an early exploration into hybrid methods. In 2021, the introduction of FAGCN \cite{fagcn_bo2021beyond} marked a significant development by incorporating adaptive edge-aware weighted techniques, which arises the flourishing exploration of adaptive message aggregation. Subsequent in 2022 and 2023, witnessed the integration of more advanced techniques with heterophilic GNNs, including dynamic routing (NCGNN \cite{yang2022ncgnn}), self-supervised learning (GREET \cite{liu2023beyond}), and attention mechanisms with memory (MMP \cite{chen2022memory}).

\vfill

\end{document}